\documentclass[letterpaper]{article} 
\usepackage{aaai23}  
\usepackage{times}  
\usepackage{helvet}  
\usepackage{courier}  
\usepackage[hyphens]{url}  
\usepackage{graphicx} 
\usepackage{natbib}  
\usepackage{caption} 
\usepackage{algorithm}
\usepackage{algorithmic}

\usepackage{newfloat}
\usepackage{listings}
\DeclareCaptionStyle{ruled}{labelfont=normalfont,labelsep=colon,strut=off} 
\floatstyle{ruled}
\newfloat{listing}{tb}{lst}{}
\floatname{listing}{Listing}
\usepackage{xspace}
\usepackage{booktabs}
\usepackage{multirow}
\usepackage{makecell}
\usepackage{xcolor}

\usepackage{amsfonts}
\usepackage{nicefrac}
\usepackage{amsmath}
\usepackage{arydshln}
\usepackage{graphicx}
\usepackage{pifont}

\usepackage{url}
\usepackage{dsfont}
\usepackage{bm}
\usepackage{pifont}
\usepackage{scalerel}
\usepackage{enumitem}
\usepackage{subcaption}
\usepackage[hang,flushmargin]{footmisc} 

\newcommand{\tb}[1]{\textbf{#1}\xspace}
\newcommand{\ul}[1]{\underline{#1}\xspace}

\newcommand{\vecb}{\boldsymbol{b}}

\newcommand{\vecp}{\boldsymbol{p}}
\newcommand{\vecq}{\boldsymbol{q}}

\newcommand{\vecu}{\boldsymbol{u}}
\newcommand{\vecv}{\boldsymbol{v}}

\newcommand{\vecx}{\boldsymbol{x}}
\newcommand{\vecy}{\boldsymbol{y}}
\newcommand{\vecz}{\boldsymbol{z}}

\newcommand{\matD}{\mathbf{D}}

\newcommand{\matH}{\mathbf{H}}

\newcommand{\matP}{\mathbf{P}}
\newcommand{\matQ}{\mathbf{Q}}

\newcommand{\matS}{\mathbf{S}}

\newcommand{\matW}{\mathbf{W}}
\newcommand{\matX}{\mathbf{X}}
\newcommand{\matY}{\mathbf{Y}}
\newcommand{\matZ}{\mathbf{Z}}

\newcommand{\calB}{\mathcal{B}}

\newcommand{\calD}{\mathcal{D}}

\newcommand{\calH}{\mathcal{H}}

\newcommand{\calL}{\mathcal{L}}

\newcommand{\calT}{\mathcal{T}}

\newcommand{\bbR}{\mathbb{R}}

\newcommand{\methodN}{TriCL-N\xspace}
\newcommand{\methodNE}{TriCL-NG\xspace}
\newcommand{\method}{TriCL\xspace}
\newcommand{\methodsampling}{TriCL-Subsampling\xspace}

\newcommand{\MLP}{MLP\xspace}
\newcommand{\GCN}{GCN\xspace}
\newcommand{\GAT}{GAT\xspace}
\newcommand{\HGNN}{HGNN\xspace}
\newcommand{\HyperConv}{HyperConv\xspace}
\newcommand{\HNHN}{HNHN\xspace}
\newcommand{\HyperGCN}{HyperGCN\xspace}
\newcommand{\HyperSAGE}{HyperSAGE\xspace}

\newcommand{\UniGCN}{UniGCN\xspace}
\newcommand{\AllSetTransformer}{AllSetTransformer\xspace}
\newcommand{\AllSet}{AllSet\xspace}

\newcommand{\Nodetovec}{Node2vec\xspace}
\newcommand{\DGI}{DGI\xspace}
\newcommand{\GRACE}{GRACE\xspace}
\newcommand{\HHGR}{S$\textsuperscript{2}$-HHGR\xspace}
\newcommand{\RandomInit}{Random-Init\xspace}

\newcommand{\weightg}{\omega_{g}}
\newcommand{\weightm}{\omega_{m}}

\newcommand{\repo}{\url{https://github.com/wooner49/TriCL}}

\title{I'm Me, We're Us, and I'm Us:\\Tri-directional Contrastive Learning on Hypergraphs}
\author{
  Dongjin~Lee$^1$ and Kijung~Shin$^{1,2}$
}
\affiliations{
  $^1$School of Electrical Engineering, $^2$Kim Jaechul Graduate School of AI, KAIST, South Korea\\
  \{dongjin.lee,\,kijungs\}@kaist.ac.kr
}

\begin{document}

\maketitle

\begin{abstract}
Although machine learning on hypergraphs has attracted considerable attention, most of the works have focused on (semi-)supervised learning, which may cause heavy labeling costs and poor generalization. 
Recently, contrastive learning has emerged as a successful unsupervised representation learning method.
Despite the prosperous development of contrastive learning in other domains, contrastive learning on hypergraphs remains little explored.
In this paper, we propose \textbf{\method} (\underline{Tri}-directional \underline{C}ontrastive \underline{L}earning), a general framework for contrastive learning on hypergraphs.
Its main idea is tri-directional contrast, and specifically, it aims to maximize in two augmented views the agreement (a) between the same node, (b) between the same group of nodes, and (c) between each group and its members. 
Together with simple but surprisingly effective data augmentation and negative sampling schemes, these three forms of contrast enable \method to capture both node- and group-level structural information in node embeddings.
Our extensive experiments using 14 baseline approaches, 10 datasets, and two tasks demonstrate the effectiveness of \method, and most noticeably, \method almost consistently outperforms not just unsupervised competitors but also (semi-)supervised competitors mostly by significant margins for node classification. 
The code and datasets are available at \repo.
\end{abstract}

\section{Introduction}
\label{sec:intro}

Many real-world interactions are group-wise. Examples include collaborations of researchers, discussions on online Q\&A sites, group conversations on messaging apps, co-citations of documents, and co-purchases of items.
A \textit{hypergraph}, which is a generalized graph, allows an edge to join an arbitrary number of nodes, and thus each such edge, which is called a \textit{hyperedge}, naturally represents such group-wise interactions~\citep{benson2018simplicial,do2020structural,lee2020hypergraph}.

Recently, machine learning on hypergraphs has drawn a lot of attention from a broad range of fields, including social network analysis~\citep{yang2019revisiting}, recommender systems~\citep{xia2021self}, and bioinformatics~\citep{zheng2019gene}.
Hypergraph-based approaches often outperform graph-based ones on various machine learning tasks, including classification \citep{feng2019hypergraph}, clustering \citep{benson2016higher}, ranking \citep{yu2021self}, and outlier detection \citep{lee2022hashnwalk}.

Previous studies have largely focused on developing encoder architectures so-called \textit{hypergraph neural networks} for hypergraph-structured data~\citep{feng2019hypergraph,yadati2019hypergcn,dong2020hnhn,bai2021hypergraph,arya2020hypersage}, and in most cases, such hypergraph neural networks are trained in a (semi-)supervised way.
However, data labeling is often time, resource, and labor-intensive, and neural networks trained only in a supervised way can easily overfit and may fail to generalize~\citep{rong2019dropedge}, making it difficult to be applied to other tasks.

Thus, self-supervised learning~\citep{liu2022graph,jaiswal2020survey,liu2021self}, which does not require labels, has become popular, and especially contrastive learning has achieved great success in computer vision~\citep{chen2020simple,hjelm2018learning} and natural language processing ~\citep{gao2021simcse}.
Contrastive learning has proved effective also for learning on (ordinary) graphs~\citep{velivckovic2018deep,peng2020graph,hassani2020contrastive,zhu2020deep,zhu2021graph,you2020graph}, and a common approach is to (a) create two augmented views from the input graph and (b) learn machine learning models to maximize the agreement between the two views.

However, contrastive learning on hypergraphs remains largely underexplored with only a handful of previous studies \citep{xia2021self,zhang2021double,yu2021self} (see Section~\ref{sec:related} for details).
Especially, the following questions remain open: (Q1) what to contrast?, (Q2) how to augment a hypergraph?, and (Q3) how to select negative samples?

For Q1, which is our main focus, we propose \textit{tri-directional contrast}. In addition to \textit{node-level contrast}, which is the only form of contrast employed in the previous studies,  we propose the use of \textit{group-level} and \textit{membership-level} contrast.
That is, in two augmented views, we aim to maximize agreements (a) between the same node, (b) between the same group of nodes, and (c) between each group and its members. 
These three forms of contrast are complementary, leading to representations that capture both node- and group-level (i.e., higher-order) relations in hypergraphs.

In addition, for Q2, we demonstrate that combining two simple augmentation strategies (spec., membership corruption and feature corruption) is effective. For Q3, we reveal that uniform random sampling is surprisingly successful, and in our experiments, even an extremely small sample size leads to marginal performance degradation.

Our proposed method \method, which is based on the aforementioned observations, is evaluated extensively using 14 baseline approaches, 10 datasets, and two tasks.
The most notable result is that, for node classification, \method outperforms not just unsupervised competitors but also all (semi-)supervised competitors on almost all considered datasets, mostly by considerable margins.
Moreover, we demonstrate the consistent effectiveness of tri-directional contrast, which is our main contribution.

\section{Related Work}
\label{sec:related}

\paragraph{Hypergraph learning}
Due to its enough expressiveness to capture higher-order structural information, learning on hypergraphs has received a lot of attention.
Many recent studies have focused on generalizing graph neural networks (GNNs) to hypergraphs~\citep{feng2019hypergraph,bai2021hypergraph,yadati2019hypergcn}.
Most of them redefine hypergraph message aggregation schemes based on clique expansion (i.e.,  replacing hyperedges with cliques to obtain a graph) or its variants.
While its simplicity is appealing, clique expansion causes structural distortion and leads to undesired information loss~\citep{hein2013total,li2018submodular}.
On the other hand, \HNHN~\citep{dong2020hnhn} prevents information loss by extending star expansion using two distinct weight matrices for node- and hyperedge-side message aggregations.
\citet{arya2020hypersage} propose \HyperSAGE for inductive learning on hypergraphs based on two-stage message aggregation.
Several studies attempt to unify hypergraphs and GNNs \citep{huang2021unignn,zhang2022hypergraph}; and
\citet{chien2021you} generalize message aggregation methods as multiset functions learned by Deep Sets~\citep{zaheer2017deep} and Set Transformer~\citep{lee2019set}.
Most approaches above use (semi-)supervised learning.

\paragraph{Contrastive learning}
In the image domain, the latest contrastive learning frameworks (e.g., SimCLR~\citep{chen2020simple} and MoCo~\citep{he2020momentum}) leverage the unchanging semantics under various image transformations, such as random flip, rotation, color distortion, etc, to learn visual features. 
They aim to learn distinguishable representations by contrasting positive and negative pairs.
 
In the graph domain, \DGI~\citep{velivckovic2018deep} combines the power of GNNs and contrastive learning, which seeks to maximize the mutual information between node embeddings and graph embeddings.
Recently, a number of graph contrastive learning approaches~\citep{you2020graph,zhu2020deep,zhu2021graph,hassani2020contrastive} that follow a common framework~\citep{chen2020simple} have been proposed.
Although these methods have achieved state-of-the-art performance on their task of interest, they cannot naturally exploit group-wise interactions, which we focus on in this paper.
More recently, gCooL~\citep{li2022graph} utilizes community contrast, which is a similar concept to membership-level contrast in \method, to maximize community consistency between two augmented views.
However, gCooL has an information loss when constructing a community, thus information on subgroups (i.e., a smaller group in a large community) cannot be used. 
On the other hand, \method can preserve and fully utilize such group information.

\paragraph{Hypergraph contrastive learning}
Contrastive learning on hypergraphs is still in its infancy.
Recently, several studies explore contrastive learning on hypergraphs~\citep{zhang2021double,xia2021self,yu2021self}.
For example, \citet{zhang2021double} proposes \HHGR for group recommendation, which applies contrastive learning to remedy a data sparsity issue.
In particular, they propose a hypergraph augmentation scheme that uses a coarse- and fine-grained node dropout for each view. However, they do not consider group-wise contrast.
Although \citet{xia2021self} employ group-wise contrast for session recommendation, they do not account for node-wise and node-group pair-wise relationships when constructing their contrastive loss.
Moreover, these approaches have been considered only in the context of group-based recommendation but not in the context of general representation learning.
\vspace{-1mm}

\section{Proposed Method: \method}
\label{sec:method}

In this section, we describe \method, our proposed framework for hypergraph contrastive learning. 
First, we introduce some preliminaries on hypergraphs and hypergraph neural networks, and then we elucidate the problem setting and details of the proposed method.

\begin{figure*}[!t]
    \vspace{-2mm}
    \centering
    \includegraphics[width=0.85\linewidth]{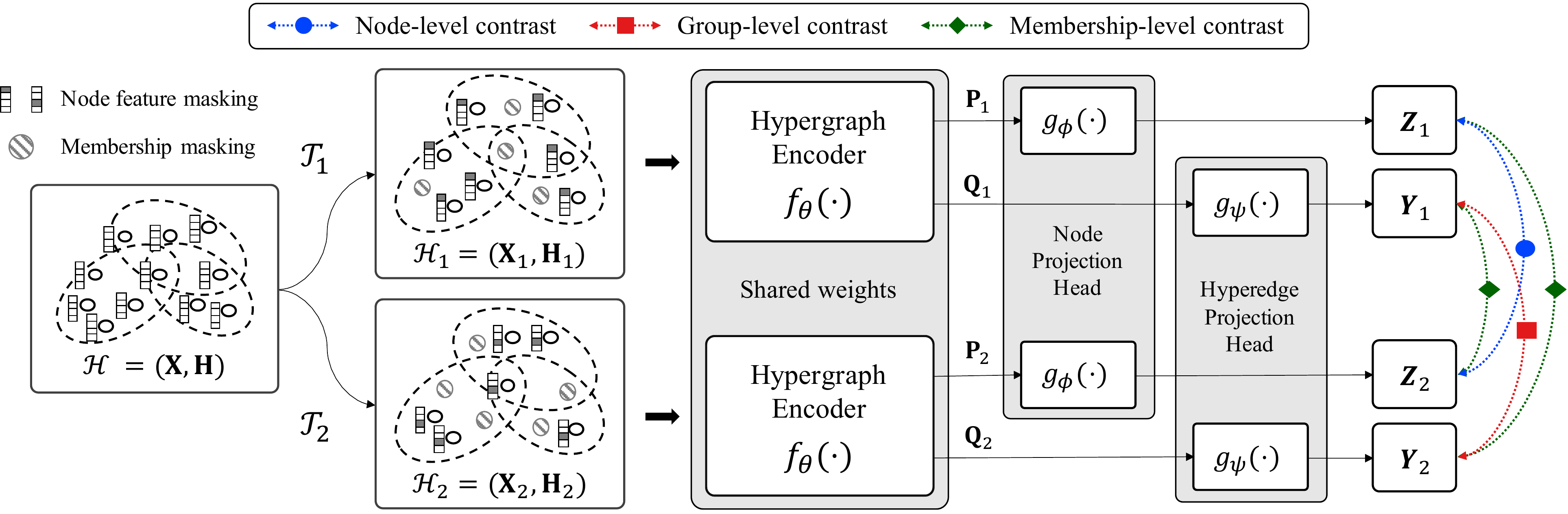}
    \centering
    \caption{ \label{fig:diagram}
        Overview of our proposed \method method.
        First, two different semantically similar views are generated by augmentations $\calT_{1}$ and $\calT_{2}$ from the original hypergraph.
        From these, we use a shared hypergraph encoder $f_{\theta}(\cdot)$ to form node and hyperedge representations. 
        After passing node and hyperedge representations to their respective projection heads (i.e., $g_{\phi}(\cdot)$ and $g_{\psi}(\cdot)$), we maximize the agreement between two views via our proposed tri-directional contrast, which is a combination of node-, group-, and membership-level contrast.
    }
     \vspace{-2mm}
\end{figure*}

\subsection{Preliminaries}
\label{sec:method:prelim}

\paragraph{Hypergraphs and notation.} 
A \textit{hypergraph}, a set of hyperedges, is a natural extension of a graph, allowing the hyperedge to contain any number of nodes.
Formally, let $H=(V,E)$ be a hypergraph, where $V=\{v_{1},v_{2},\dots,v_{|V|}\}$ is a set of nodes and $E=\{e_{1},e_{2},\dots,e_{|E|}\}$ is a set of hyperedges, with each hyperedge is a non-empty subset of $V$.
The node feature matrix is represented by $\matX\in\bbR^{|V|\times F}$, where $\vecx_{i}=\matX[i,:]^{T}\in\bbR^{F}$ is the feature of node $v_{i}$.
In general, a hypergraph can alternatively be represented by its \textit{incidence matrix} $\matH\in\{0,1\}^{|V|\times |E|}$, with entries defined as $h_{ij}=1$ if $v_{i}\in e_{j}$, and $h_{ij}=0$ otherwise. 
In other words, $h_{ij}=1$ when node $v_{i}$ and hyperedge $e_{j}$ form a \textit{membership}.
Each hyperedge $e_{j}\in E$ is assigned a positive weight $w_{j}$, and all the weights formulate a diagonal matrix $\matW\in\bbR^{|E|\times |E|}$.
We use the diagonal matrix $\matD_{V}$ to represent the degree of vertices, where its entries $d_{i}=\sum_{j}{w_{j}h_{ij}}$.
Also we use the diagonal matrix $\matD_{E}$ to denote the degree of hyperedges, where its element $\delta_{j}=\sum_{i}{h_{ij}}$ represents the number of nodes connected by the hyperedge $e_{j}$.

\paragraph{Hypergraph neural networks.}
Modern hypergraph neural networks~\citep{feng2019hypergraph,yadati2019hypergcn,bai2021hypergraph,dong2020hnhn,arya2020hypersage,chien2021you} follow a two-stage neighborhood aggregation strategy: node-to-hyperedge and hyperedge-to-node aggregation. 
They iteratively update the representation of a hyperedge by aggregating representations of its incident nodes and the representation of a node by aggregating representations of its incident hyperedges.
Let $\matP^{(k)}\in\bbR^{|V|\times F_{k}'}$ and $\matQ^{(k)}\in\bbR^{|E|\times F_{k}''}$ be the node and hyperedge representations at the $k$-th layer, respectively.
Formally, the $k$-th layer of a hypergraph neural network is 
\begin{equation}\label{eq:hypergnn}
\begin{split}
    & \vecq_{j}^{(k)}=f_{V\rightarrow E}^{(k)}\Big(\vecq_{j}^{(k-1)},\big\{\vecp_{i}^{(k-1)}:v_{i}\in e_{j}\big\}\Big),\\
    & \vecp_{i}^{(k)}=f_{E\rightarrow V}^{(k)}\Big(\vecp_{i}^{(k-1)},\big\{\vecq_{j}^{(k)}:v_{i}\in e_{j}\big\}\Big),
\end{split}
\end{equation}
where $\vecp_{i}^{(0)}=\vecx_{i}$.
The choice of aggregation rules, $f_{V\rightarrow E}(\cdot)$ and $f_{E\rightarrow V}(\cdot)$, is critical, and a number of models have been proposed.
In HGNN~\citep{feng2019hypergraph}, for example, they choose $f_{V\rightarrow E}$ and $f_{E\rightarrow V}$ to be the weighted sum over inputs with normalization as:
\small
\begin{equation}\label{eq:hgnn}
    \vecq_{j}^{(k)}=\sum_{v_{i}\in e_{j}}{\frac{\vecp_{i}^{(k-1)}}{\sqrt{d_{i}}}},\ 
    \vecp_{i}^{(k)}=\sigma\bigg(\frac{1}{\sqrt{d_{i}}}\sum_{e_{j}:v_{i}\in e_{j}}\frac{w_{j}\vecq_{j}^{(k)}\mathbf{\Theta}^{(k)}}{\delta_{j}}+\vecb^{(k)}\bigg),
\end{equation}
\normalsize
where $\mathbf{\Theta}^{(k)}$ is a learnable weight matrix, $\vecb^{(k)}$ is a bias, and $\sigma$ denotes a non-linear activation function.
Many other hypergraph neural networks can be represented by (\ref{eq:hypergnn}).

\subsection{Problem Setting: Hypergraph-based Contrastive Learning}
\label{sec:method:problem}
Our objective is to train a hypergraph encoder, $f_{\theta}:\bbR^{|V|\times F}\times\bbR^{|V|\times|E|}\rightarrow\bbR^{|V|\times F'}\times\bbR^{|E|\times F''}$, such that $f_{\theta}(\matX,\matH)=(\matP,\matQ)$ produces low-dimensional representations of nodes and hyperedges in a fully unsupervised manner, specifically a contrastive manner.
These representations may then be utilized for downstream tasks, such as node classification and clustering.

\subsection{\method: Tri-directional Contrastive Learning}
\label{sec:method:proposed}

Basically, \method follows the conventional multi-view graph contrastive learning paradigm, where a model aims to maximize the agreement of representations between different views~\citep{you2020graph,hassani2020contrastive,zhu2020deep}.
While most existing approaches only use node-level contrast, \method applies three forms of contrast for each of the three essential elements constituting hypergraphs: nodes, hyperedges, and node-hyperedge memberships.
Figure~\ref{fig:diagram} visually summarizes \method's architecture.
\method is composed of the following four major components:

\paragraph{(1) Hypergraph augmentation.}
We consider a hypergraph $\calH=(\matX,\matH)$.
\method first generates two alternate views of the hypergraph $\calH$: $\calH_{1}=(\matX_{1},\matH_{1})$ and $\calH_{2}=(\matX_{2},\matH_{2})$, by applying stochastic hypergraph augmentation function $\calT_{1}$ and $\calT_{2}$, respectively.
We use a combination of random \textit{node feature masking}~\citep{you2020graph,zhu2020deep} and \textit{membership masking} to augment a hypergraph in terms of attributes and structure.
Following previous studies~\citep{you2020graph,thakoor2021large}, node feature masking is not applied to each node independently, and instead, we generate a single random binary mask of size $F$ where each entry is sampled from a Bernoulli distribution $\calB(1-p_{f})$, and we use it to mask features of all nodes.
Similarly, we use a binary mask of size $K=nnz(\matH)$ where each element is sampled from a Bernoulli distribution $\calB(1-p_{m})$ to mask node-hyperedge memberships.
The degree of augmentation is controlled by $p_{f}$ and $p_{m}$, and we can adopt different hyperparameters for each augmented view.
More details on hypergraph augmentation are provided in Appendix~\ref{sec:app:augmentation}.

\paragraph{(2) Hypergraph encoder.}
A hypergraph encoder $f_{\theta}(\cdot)$ produces node and hyperedge representations, $\matP$ and $\matQ$, respectively, for two augmented views: $(\matP_{1}, \matQ_{1}):=f_{\theta}(\matX_{1},\matH_{1})$ and $(\matP_{2},\matQ_{2}):=f_{\theta}(\matX_{2},\matH_{2})$.
\method does not constrain the choice of hypergraph encoder architectures if they can be formulated by (\ref{eq:hypergnn}). In our proposed method, we use the element-wise mean pooling layer as a special instance of (\ref{eq:hypergnn}) (see Appendix~\ref{sec:app:exp:encoder} for comparison with an alternative).
That is, $f_{V\rightarrow E}$ and $f_{E\rightarrow V}$ are as:
\small
\begin{equation}\label{eq:encoder}
\begin{split}
& \vecq_{j}^{(k)}=\sigma\bigg(\sum_{v_{i}\in e_{j}}{\frac{\vecp_{i}^{(k-1)}\mathbf{\Theta}_{E}^{(k)}}{\delta_{j}}}+\vecb_{E}^{(k)}\bigg),\\
& \vecp_{i}^{(k)}=\sigma\bigg(\sum_{e_{j}:v_{i}\in e_{j}}{\frac{w_{j}\vecq_{j}^{(k)}\mathbf{\Theta}_{V}^{(k)}}{d_{i}}}+\vecb_{V}^{(k)}\bigg),
\end{split}
\end{equation}
\normalsize
where $\mathbf{\Theta}_{E}^{(k)}$ and $\mathbf{\Theta}_{V}^{(k)}$ are trainable weights and $\vecb_{E}^{(k)}$ and $\vecb_{V}^{(k)}$ are trainable biases.
We use $w_{j}=1$ for simplicity, and
(\ref{eq:encoder}) can be represented as (\ref{eq:encoder:matrix}) in matrix form.
\begin{equation}\label{eq:encoder:matrix}
\begin{split}
& \matQ^{(k)}=\sigma\big(\matD_{E}^{-1}\matH^{T}\matP^{(k-1)}\mathbf{\Theta}_{E}^{(k)}+\vecb_{E}^{(k)}\big),\\
& \matP^{(k)}=\sigma\big(\matD_{V}^{-1}\matH\matW\matQ^{(k)}\mathbf{\Theta}_{V}^{(k)}+\vecb_{V}^{(k)}\big),
\end{split}
\end{equation}
where $\matP^{(0)}=\matX$ and $\matW$ is the identity matrix.

\paragraph{(3) Projection head.}
\citet{chen2020simple} empirically demonstrate that including a non-linear transformation called projection head which maps representations to another latent space where contrastive loss is applied helps to improve the quality of representations.
We also adopt two projection heads denoted by $g_{\phi}(\cdot)$ and $g_{\psi}(\cdot)$ for projecting node and hyperedge representations, respectively.
Both projection heads in our method are implemented with a two-layer MLP and ELU activation~\citep{clevert2015fast}.
Formally, $\matZ_{k}:=g_{\phi}(\matP_{k})$ and $\matY_{k}:=g_{\psi}(\matQ_{k})$, where $k=1,2$ for two augmented views.

\paragraph{(4) Tri-directional contrastive loss.}
In \method framework, we employ three contrastive objectives:
(a) \textbf{node-level contrast} aims to discriminate the representations of the same node in the two augmented views from other node representations, 
(b) \textbf{group-level contrast} tries to distinguish the representations of the same hyperedge in the two augmented views from other hyperedge representations, and 
(c) \textbf{membership-level contrast} seeks to differentiate a ``real'' node-hyperedge membership from a ``fake'' one across the two augmented views.
We utilize the InfoNCE loss~\citep{oord2018representation}, one of the popular contrastive losses, as in 
\citep{zhu2020deep,zhu2021graph,qiu2020gcc}.

In the rest of this subsection, we first provide a motivating example for the tri-directional contrastive loss. Then, we describe each of its three components in detail.

\paragraph{Motivating example.}
\textit{How can the three forms of contrast be helpful for node representation learning?}
In node classification tasks, for example, information about a group of nodes could help improve performance. 
Specifically, in co-authorship networks such as Cora-A and DBLP, nodes and hyperedges represent papers and authors, respectively, and papers written by the same author are more likely to belong to the same field and cover similar topics (i.e. homophily exists in hypergraphs~\citep{veldt2021higher}). 
Thus, high-quality author information could be useful in inferring the field of the papers he wrote, especially when information about the paper is insufficient.

Furthermore, leveraging node-hyperedge membership helps enrich the information of each node and hyperedge. 
For example, the fact that a meteorology paper is written by an author who studies mainly machine learning is a useful clue to suspect that (a) the paper is about application of machine learning techniques to meteorological problems and (b) the author is interested not only in machine learning but also in meteorology. 
In order to utilize such benefits explicitly, we proposed the tri-directional contrastive loss, which is described below.

\paragraph{Node-level contrast.}
For any node $v_{i}$, its representation from the first view, $\vecz_{1,i}$, is set to the anchor, the representation of it from the second view, $\vecz_{2,i}$, is treated as the positive sample, and the other representations from the second view, $\vecz_{2,k}$, where $k\neq i$, are regarded as negative samples.
Let $s(\cdot,\cdot)$ denote the \textit{score} function (a.k.a. \textit{critic} function) that assigns high values to the positive pair, and low values to negative pairs~\citep{tschannen2019mutual}.
We use the cosine similarity as the score (i.e. $s(\vecu,\vecv)=\vecu^{T}\vecv/\|\vecu\|\|\vecv\|$).
Then the loss function for each positive node pair is defined as:
\begin{equation*}\label{eq:loss:node}
\ell_{n}(\vecz_{1,i},\vecz_{2,i})
=-\log{\frac{e^{s(\vecz_{1,i},\vecz_{2,i})/\tau_{n}}}{\sum_{k=1}^{|V|}{e^{s(\vecz_{1,i},\vecz_{2,k})/\tau_{n}}}}},
\end{equation*}
where $\tau_{n}$ is a temperature parameter.  In practice, we symmetrize this loss by setting the node representation of the second view as the anchor. 
The objective function for node-level contrast is the average over all positive pairs as:
\begin{equation}\label{eq:loss:node:sum}
\calL_{n}=\frac{1}{2|V|}\sum_{i=1}^{|V|}{\big\{\ell_{n}(\vecz_{1,i},\vecz_{2,i}) + \ell_{n}(\vecz_{2,i},\vecz_{1,i})\big\}}.
\end{equation}

\paragraph{Group-level contrast.}
For any hyperedge (i.e., a group of nodes) $e_{j}$, its representation from the first view, $\vecy_{1,j}$, is set to the anchor, the representation of it from the other view, $\vecy_{2,j}$, is treated as the positive sample, and the other representations from the view where the positive samples lie, $\vecy_{2,k}$, where $k\neq j$, are regarded as negative samples.
We also use the cosine similarity as the critic,
and then the loss function for each positive hyperedge pair is defined as:
\begin{equation*}\label{eq:loss:hyperedge}
\ell_{g}(\vecy_{1,j},\vecy_{2,j})
=-\log{\frac{e^{s(\vecy_{1,j},\vecy_{2,j})/\tau_{g}}}{\sum_{k=1}^{|E|}{e^{s(\vecy_{1,j},\vecy_{2,k})/\tau_{g}}}}},
\end{equation*}
where $\tau_{g}$ is a temperature parameter. 
The objective function for group-level contrast is defined as: 
\begin{equation}\label{eq:loss:hyperedge:sum}
\calL_{g}=\frac{1}{2|E|}\sum_{j=1}^{|E|}{\big\{\ell_{g}(\vecy_{1,j},\vecy_{2,j}) + \ell_{g}(\vecy_{2,j},\vecy_{1,j})\big\}}.
\end{equation}

\paragraph{Membership-level contrast.}
For any node $v_{i}$ and hyperedge $e_{j}$ that form membership (i.e., $v_i\in e_j$) in the original hypergraph, the node representation from the first view, $\vecz_{1,i}$, is set to the anchor, the hyperedge representation from the other view, $\vecy_{2,j}$, is treated as the positive sample.
The negative samples are drawn from the representations of the other hyperedges
that are not associated with node $v_{i}$, denoted by $\vecy_{2,k}$, where $k:i\notin k$.
Symmetrically, $\vecy_{2,j}$ can also be the anchor, in which case the negative samples are $\vecz_{1,k}$, where $k:k\notin j$.
To differentiate a ``real'' node-hyperedge membership from a ``fake'' one, 
we employ a \textit{discriminator}, $\calD:\bbR^{F'}\times\bbR^{F''}\rightarrow\bbR$ as the scoring function so that $\calD(\vecz,\vecy)$ represents the probability scores assigned to this node-hyperedge representation pair (should be higher for ``real'' pairs)~\citep{hjelm2018learning,velivckovic2018deep}.
For simplicity, we omit the augmented view number in the equation.
Then we use the following objective:
\small
\begin{equation*}\label{eq:loss:membership}
\begin{split}
\ell_{m}(\vecz_{i},\vecy_{j})=&-\underbrace{\log{\frac{e^{\calD(\vecz_{i},\vecy_{j})/\tau_{m}}}{e^{\calD(\vecz_{i},\vecy_{j})/\tau_{m}}+\sum_{k:i\notin k}{e^{\calD(\vecz_{i},\vecy_{k})/\tau_{m}}}}}}_{{\textstyle{\scriptsize\text{when $\vecz_{i}$ is the anchor}}}}\\
&-\underbrace{\log{\frac{e^{\calD(\vecz_{i},\vecy_{j})/\tau_{m}}}{e^{\calD(\vecz_{i},\vecy_{j})/\tau_{m}}+\sum_{k:k\notin j}{e^{\calD(\vecz_{k},\vecy_{j})/\tau_{m}}}}}}_{\textstyle{\scriptsize\text{when $\vecy_{j}$ is the anchor}}},
\end{split}
\end{equation*}
\normalsize
where $\tau_{m}$ is a temperature parameter.
From a practical point of view, considering a large number of negatives poses a prohibitive cost, especially for large graphs~\citep{zhu2020deep,thakoor2021large}. 
We, therefore, decide to randomly select a single negative sample per positive sample for $\ell_{m}(\vecz_{i},\vecy_{j})$.
Since two views are symmetric, we get two node-hyperedge pairs for a single membership.
The objective function for membership-level contrast is defined as:
\small
\begin{equation}\label{eq:loss:membership:sum}
\calL_{m}=\frac{1}{2K}\sum_{i=1}^{|V|}{\sum_{j=1}^{|E|}{\mathds{1}_{[h_{ij}=1]}\big\{\ell_{m}(\vecz_{1,i},\vecy_{2,j})+\ell_{m}(\vecz_{2,i},\vecy_{1,j})\big\}}}.
\end{equation}
\normalsize

Finally, by integrating Eq. (\ref{eq:loss:node:sum}), (\ref{eq:loss:hyperedge:sum}), and (\ref{eq:loss:membership:sum}), our proposed contrastive loss is formulated as:
\begin{equation}\label{eq:loss:final}
\calL=\calL_{n}+\weightg\calL_{g}+\weightm\calL_{m},
\end{equation}
\mbox{where $\weightg$ and $\weightm$ are the weights of $\calL_{g}$ and $\calL_{m}$, respectively.}

To sum up, \method jointly optimizes three contrastive objectives (i.e., node-, group-, and membership-level contrast), which enable the learned embeddings of nodes and hyperedges to preserve both the node- and group-level structural information at the same time.

\begin{table*}[!t]
    \caption{Node classification accuracy and standard deviations. 
    Graph methods, marked as $\star$, are applied after converting hypergraphs to graphs via clique expansion.
    For each dataset, the best and the second-best performances are highlighted in \tb{boldface} and \ul{underlined}, respectively.
    A.R. denotes average rank, OOT denotes cases where results are not obtained within 24 hours, and OOM indicates out of memory on a 24GB GPU.
    In most cases, \method outperforms all others, including the supervised ones.
    }
    \label{tab:accuracy}
    \centering
    \scalebox{0.693}{
        \begin{tabular}{clccccccccccc}
            \toprule
            & Method & Cora-C & Citeseer & Pubmed & Cora-A & DBLP & Zoo & 20News & Mushroom & NTU2012 & ModelNet40 & A.R.$\downarrow$ \\
            \midrule
            \midrule
            \multirow{10}{*}{\rotatebox{90}{Supervised}} 
                & \MLP              & 60.32 $\pm$ 1.5   & 62.06 $\pm$ 2.3  & 76.27 $\pm$ 1.1        & 64.05 $\pm$ 1.4   & 81.18 $\pm$ 0.2   & 75.62 $\pm$ 9.5   & 79.19 $\pm$ 0.5       & 99.58 $\pm$ 0.3   & 65.17 $\pm$ 2.3   & 93.75 $\pm$ 0.6  & 12.5   \\
                & \GCN\!$^{\star}$  & 77.11 $\pm$ 1.8   & 66.07 $\pm$ 2.4  & 82.63 $\pm$ 0.6        & 73.66 $\pm$ 1.3   & 87.58 $\pm$ 0.2   & 36.79 $\pm$ 9.6   & OOM                   & 92.47 $\pm$ 0.9   & 71.17 $\pm$ 2.4   & 91.67 $\pm$ 0.2  & 11.7   \\
                & \GAT\!$^{\star}$  & 77.75 $\pm$ 2.1   & 67.62 $\pm$ 2.5  & 81.96 $\pm$ 0.7        & 74.52 $\pm$ 1.3   & 88.59 $\pm$ 0.1   & 36.48 $\pm$ 10.0  & OOM                   & OOM               & 70.94 $\pm$ 2.6   & 91.43 $\pm$ 0.3  & 11     \\
                & \HGNN             & 77.50 $\pm$ 1.8   & 66.16 $\pm$ 2.3  & 83.52 $\pm$ 0.7        & 74.38 $\pm$ 1.2   & 88.32 $\pm$ 0.3   & 78.58 $\pm$ 11.1  & 80.15 $\pm$ 0.3       & 98.59 $\pm$ 0.5   & 72.03 $\pm$ 2.4   & 92.23 $\pm$ 0.2  & 8.1    \\
                & \HyperConv        & 76.19 $\pm$ 2.1   & 64.12 $\pm$ 2.6  & 83.42 $\pm$ 0.6        & 73.52 $\pm$ 1.0   & 88.83 $\pm$ 0.2   & 62.53 $\pm$ 14.5  & 79.83 $\pm$ 0.4       & 97.56 $\pm$ 0.6   & 72.62 $\pm$ 2.6   & 91.84 $\pm$ 0.1  & 9.8    \\
                & \HNHN             & 76.21 $\pm$ 1.7   & 67.28 $\pm$ 2.2  & 80.97 $\pm$ 0.9        & 74.88 $\pm$ 1.6   & 86.71 $\pm$ 1.2   & 78.89 $\pm$ 10.2  & 79.51 $\pm$ 0.4       & 99.78 $\pm$ 0.1   & 71.45 $\pm$ 3.2   & 92.96 $\pm$ 0.2  & 8.9    \\
                & \HyperGCN         & 64.11 $\pm$ 7.4   & 59.92 $\pm$ 9.6  & 78.40 $\pm$ 9.2        & 60.65 $\pm$ 9.2   & 76.59 $\pm$ 7.6   & 40.86 $\pm$ 2.1   & 77.31 $\pm$ 6.0       & 48.26 $\pm$ 0.3   & 46.05 $\pm$ 3.9   & 69.23 $\pm$ 2.8  & 15.1   \\
                & \HyperSAGE        & 64.98 $\pm$ 5.3   & 52.43 $\pm$ 9.4  & 79.49 $\pm$ 8.7        & 64.59 $\pm$ 4.3   & 79.63 $\pm$ 8.6   & 40.86 $\pm$ 2.1   & OOT              & OOT          & OOT          & OOT         & 14.7   \\
                & \UniGCN           & 77.91 $\pm$ 1.9   & 66.40 $\pm$ 1.9  & \ul{84.08 $\pm$ 0.7}   & 77.30 $\pm$ 1.4   & 90.31 $\pm$ 0.2   & 72.10 $\pm$ 12.1  & \tb{80.24 $\pm$ 0.4}  & 98.84 $\pm$ 0.5   & 73.27 $\pm$ 2.7   & 94.62 $\pm$ 0.2  & 5.9    \\
                & \AllSet           & 76.21 $\pm$ 1.7   & 67.83 $\pm$ 1.8  & 82.85 $\pm$ 0.9        & 76.94 $\pm$ 1.3   & 90.07 $\pm$ 0.3   & 72.72 $\pm$ 11.8  & 79.90 $\pm$ 0.4       & 99.78 $\pm$ 0.1   & 75.09 $\pm$ 2.5   & 96.85 $\pm$ 0.2  & 6.2    \\
            \midrule
            \multirow{8}{*}{\rotatebox{90}{Unsupervised}}
                & \Nodetovec\!$^{\star}$    & 70.99 $\pm$ 1.4   & 53.85 $\pm$ 1.9   & 78.75 $\pm$ 0.9   & 58.50 $\pm$ 2.1   & 72.09 $\pm$ 0.3   & 17.02 $\pm$ 4.1   & 63.35 $\pm$ 1.7   & 88.16 $\pm$ 0.8   & 67.72 $\pm$ 2.1   & 84.94 $\pm$ 0.4   & 15.6  \\
                & \DGI\!$^{\star}$          & 78.17 $\pm$ 1.4   & 68.81 $\pm$ 1.8   & 80.83 $\pm$ 0.6   & 76.94 $\pm$ 1.1   & 88.00 $\pm$ 0.2   & 36.54 $\pm$ 9.7   & OOM               & OOM               & 72.01 $\pm$ 2.5   & 92.18 $\pm$ 0.2   & 9.3   \\
                & \GRACE\!$^{\star}$        & 79.11 $\pm$ 1.7   & 68.65 $\pm$ 1.7   & 80.08 $\pm$ 0.7   & 76.59 $\pm$ 1.0   & OOM               & 37.07 $\pm$ 9.3   & OOM               & OOM               & 70.51 $\pm$ 2.4   & 90.68 $\pm$ 0.3   & 10.4  \\
                & \HHGR                     & 78.08 $\pm$ 1.7   & 68.21 $\pm$ 1.8   & 82.13 $\pm$ 0.6   & 78.15 $\pm$ 1.1   & 88.69 $\pm$ 0.2   & 80.06 $\pm$ 11.1  & 79.75 $\pm$ 0.3   & 97.15 $\pm$ 0.5   & 73.95 $\pm$ 2.4   & 93.26 $\pm$ 0.2   & 6.8   \\
                \cmidrule{2-13}
                & \RandomInit           & 63.62 $\pm$ 3.1       & 60.44 $\pm$ 2.5       & 67.49 $\pm$ 2.2       & 66.27 $\pm$ 2.2       & 76.57 $\pm$ 0.6       & 78.43 $\pm$ 11.0      & 77.14 $\pm$ 0.6       & 97.40 $\pm$ 0.6       & 74.39 $\pm$ 2.6       & 96.29 $\pm$ 0.3       & 11.9      \\
                & \tb{\methodN}         & 80.23 $\pm$ 1.2       & 70.28 $\pm$ 1.5       & 83.44 $\pm$ 0.6       & 81.94 $\pm$ 1.1       & 90.88 $\pm$ 0.1       & 79.94 $\pm$ 11.1      & \ul{80.18 $\pm$ 0.2}  & 99.76 $\pm$ 0.2       & 75.20 $\pm$ 2.6       & 97.01 $\pm$ 0.2       & 3.4       \\
                & \tb{\methodNE}        & \ul{81.45 $\pm$ 1.2}  & \ul{71.38 $\pm$ 1.2}  & 83.68 $\pm$ 0.7       & \ul{82.00 $\pm$ 1.0}  & \ul{90.94 $\pm$ 0.1}  & \ul{80.19 $\pm$ 11.1} & \ul{80.18 $\pm$ 0.2}  & \ul{99.81 $\pm$ 0.1}  & \tb{75.25 $\pm$ 2.5}  & \ul{97.02 $\pm$ 0.1}  & \ul{2}    \\
                & \tb{\method}          & \tb{81.57 $\pm$ 1.1}  & \tb{72.02 $\pm$ 1.2}  & \tb{84.26 $\pm$ 0.6}  & \tb{82.15 $\pm$ 0.9}  & \tb{91.12 $\pm$ 0.1}  & \tb{80.25 $\pm$ 11.2} & 80.14 $\pm$ 0.2       & \tb{99.83 $\pm$ 0.1}  & \ul{75.23 $\pm$ 2.4}  & \tb{97.08 $\pm$ 0.1}  & \tb{1.5}  \\
            \bottomrule
        \end{tabular}
    }
\end{table*}

\section{Experiments}
\label{sec:exp}
In this section, we empirically evaluate the quality of node representations learnt by \method on two hypergraph learning tasks: node classification and clustering, which have been commonly used to benchmark hypergraph learning algorithms~\citep{zhou2006learning}.

\subsection{Dataset}
\label{sec:exp:dataset}
We assess the performance of \method on 10 commonly used benchmark datasets;
these datasets are categorized into
(1) co-citation datasets (Cora, Citeseer, and Pubmed)~\citep{sen2008collective},
(2) co-authorship datasets (Cora and DBLP~\citep{rossi2015network}), 
(3) computer vision and graphics datasets (NTU2012~\citep{chen2003visual} and ModelNet40~\citep{wu20153d}), and
(4) datasets from the UCI Categorical Machine Learning Repository~\citep{Dua:2019} (Zoo, 20Newsgroups, and Mushroom).
Further descriptions and the statistics of datasets are provided in Appendix~\ref{sec:app:dataset}.

\subsection{Experimental Setup}
\label{sec:exp:setup}

\paragraph{Evaluation protocol.}
For the node classification task, we follow the standard linear-evaluation protocol as introduced in~\citet{velivckovic2018deep}.
The encoder is firstly trained in a fully unsupervised manner and computes node representations; then, a simple linear classifier is trained on top of these frozen representations through a $\ell_{2}$-regularized logistic regression loss, without flowing any gradients back to the encoder.
For all the datasets, we randomly split them, where 10\%, 10\%, and 80\% of nodes are chosen for the training, validation, and test set, respectively, as has been followed in \citet{zhu2020deep,thakoor2021large}. 
We evaluate the model with 20 dataset splits over 5 random weight initializations for unsupervised setting, and report the averaged accuracy on each dataset.
In a supervised setting, we use 20 dataset splits and a different model initialization for each split and report the averaged accuracy. 

For the clustering task, we assess the quality of representations using the k-means clustering by operating it on the frozen node representations produced by each model.
We employ the local Lloyd algorithm~\citep{lloyd1982least} with the k-means++ seeding~\citep{arthur2006k} approach.
For a fair comparison, we train each model with 5 random weight initializations, perform k-means 5 times on each trained encoder, and report the averaged results.

\paragraph{Baselines.}
We compare \method with various representative baseline approaches including 10 (semi-)supervised models and 4 unsupervised models.
A detailed description of these baselines is provided in Appendix~\ref{sec:app:baseline}.
Note that, since the methods working on graphs can not be directly applied to hypergraphs, we use them after transforming hypergraphs to graphs via clique expansion.
For all baselines, we report their performance based on their official implementations.


\paragraph{Implementation details.}
We employ a one-layer mean pooling hypergraph encoder described in (\ref{eq:encoder:matrix}) and PReLU~\citep{he2015delving} activation for non-linearlity.
Following \citet{tschannen2019mutual}, which has experimentally shown a bilinear critic yields better downstream performance than higher-capacity MLP critics, we use a bilinear function as a discriminator to score node-hyperedge representation pairs, formulated as $\calD(\vecz,\vecy)=\sigma(\vecz^{T}\matS\vecy)$.
Here, $\matS$ denotes a trainable scoring matrix and $\sigma$ is the sigmoid function to transform scores into probabilities of $(\vecz,\vecy)$ being a positive sample.
A description of the optimizer and model hyperparameters are provided in Appendix~\ref{sec:app:impl}.

\subsection{Performance on Node Classification}
\label{sec:exp:classification}

\begin{table*}
    \caption{Comparison of node classification accuracy according to whether or not to use each type of contrast (i.e., $\calL_{n}$, $\calL_{g}$, and $\calL_{m}$). 
    Using all types of contrasts (i.e., node-, group-, and membership-level contrast) achieves the best performance in most cases as they are complementarily reinforcing each other.
    }
    \label{tab:comparison}
    \centering
    \scalebox{0.69}{
        \begin{tabular}{cccccccccccccc}
            \toprule
            $\calL_{n}$ & $\calL_{g}$ & $\calL_{m}$ & Cora-C & Citeseer & Pubmed & Cora-A & DBLP & Zoo & 20News & Mushroom & NTU2012 & ModelNet40 & A.R.$\downarrow$ \\
            \midrule
            \midrule
            \ding{51}   & -         & -         & 80.23 $\pm$ 1.2       & 70.28 $\pm$ 1.5*      & 83.44 $\pm$ 0.6       & 81.94 $\pm$ 1.1       & 90.88 $\pm$ 0.1       & 79.94 $\pm$ 11.1      & \tb{80.18 $\pm$ 0.2}  & 99.76 $\pm$ 0.2       & 75.20 $\pm$ 2.6       & 97.01 $\pm$ 0.2       & 3.8       \\
            -           & \ding{51} & -         & 79.69 $\pm$ 1.6       & 71.02 $\pm$ 1.3*      & 80.20 $\pm$ 1.3       & 78.98 $\pm$ 1.4       & 88.60 $\pm$ 0.2       & 79.31 $\pm$ 10.7      & 79.35 $\pm$ 0.4       & 99.13 $\pm$ 0.3       & 74.41 $\pm$ 2.6       & 96.66 $\pm$ 0.2       & 5.7       \\
            -           & -         & \ding{51} & 76.76 $\pm$ 1.8       & 63.98 $\pm$ 2.0       & 79.86 $\pm$ 0.9       & 76.77 $\pm$ 1.1       & 63.95 $\pm$ 7.2       & 79.80 $\pm$ 11.0      & 79.27 $\pm$ 0.3       & 94.87 $\pm$ 0.7       & 73.11 $\pm$ 2.8       & 96.57 $\pm$ 0.2       & 6.9       \\
            \ding{51}   & \ding{51} & -         & \ul{81.45 $\pm$ 1.2}  & 71.38 $\pm$ 1.4       & 83.68 $\pm$ 0.7       & \ul{82.00 $\pm$ 1.0}  & \ul{90.94 $\pm$ 0.1}  & 80.19 $\pm$ 11.1      & \tb{80.18 $\pm$ 0.2}  & \ul{99.81 $\pm$ 0.1}  & \tb{75.25 $\pm$ 2.5}  & 97.02 $\pm$ 0.1       & \ul{2.3}  \\
            \ding{51}   & -         & \ding{51} & 80.49 $\pm$ 1.3       & 70.46 $\pm$ 1.5       & \ul{83.98 $\pm$ 0.7}  & 81.62 $\pm$ 1.0       & 90.75 $\pm$ 0.1       & 80.19 $\pm$ 11.1      & 80.15 $\pm$ 0.2       & 99.74 $\pm$ 0.2       & 75.12 $\pm$ 2.5       & \ul{97.03 $\pm$ 0.1}  & 3.6       \\
            -           & \ding{51} & \ding{51} & 80.80 $\pm$ 1.1       & \ul{71.73 $\pm$ 1.4}  & 82.81 $\pm$ 0.7       & 80.24 $\pm$ 1.0       & 90.17 $\pm$ 0.1       & \ul{80.20 $\pm$ 11.1} & 79.29 $\pm$ 0.2       & 99.82 $\pm$ 0.1       & 73.76 $\pm$ 2.5       & 96.74 $\pm$ 0.1       & 4.1       \\
            \ding{51}   & \ding{51} & \ding{51} & \tb{81.57 $\pm$ 1.1}  & \tb{72.02 $\pm$ 1.4}  & \tb{84.26 $\pm$ 0.6}  & \tb{82.15 $\pm$ 0.9}  & \tb{91.12 $\pm$ 0.1}  & \tb{80.25 $\pm$ 11.2} & 80.14 $\pm$ 0.2       & \tb{99.83 $\pm$ 0.1}  & \ul{75.23 $\pm$ 2.4}  & \tb{97.08 $\pm$ 0.1}  & \tb{1.4}  \\
            \bottomrule
        \end{tabular}
    }
\end{table*}

\begin{table*}
    
    \setlength{\tabcolsep}{5pt}
    \caption{
    \method is very robust to the number of negative samples (i.e., $k$).
    A.P.D. stands for average performance degradation.
    Even if only two negative samples are used at every gradient step, the performance degrades by less than 1\%.
    }
    \label{tab:subsampling}
    \centering
    \scalebox{0.66}{
        \begin{tabular}{lccccccccccc}
            \toprule
            Method & Cora-C & Citeseer & Pubmed & Cora-A & DBLP & Zoo & 20News & Mushroom & NTU2012 & ModelNet40 & A.P.D. \\
            \midrule
            \midrule
            \HHGR all negatives     & 78.08 $\pm$ 1.7          & 68.21 $\pm$ 1.8          & 82.13 $\pm$ 0.6          & 78.15 $\pm$ 1.1          & 88.69 $\pm$ 0.2          & 80.06 $\pm$ 11.1              & 79.75 $\pm$ 0.3   & 97.15 $\pm$ 0.5   & 73.95 $\pm$ 2.4   & 93.26 $\pm$ 0.2 & -         \\
            \midrule
            \methodsampling $(k=2)$ & 80.62 $\pm$ 1.3       & 71.95 $\pm$ 1.3       & 83.22 $\pm$ 0.7       & 81.25 $\pm$ 1.0       & 90.66 $\pm$ 0.2       & 80.10 $\pm$ 11.1      & 80.03 $\pm$ 0.2       & 99.82 $\pm$ 0.1       & 74.95 $\pm$ 2.6       & 97.02 $\pm$ 0.1       & 0.49\% \\
            \methodsampling $(k=4)$ & 81.15 $\pm$ 1.2       & \tb{72.24 $\pm$ 1.2}  & 83.91 $\pm$ 0.7       & 81.85 $\pm$ 0.9       & 90.83 $\pm$ 0.1       & 80.16 $\pm$ 11.3      & 80.08 $\pm$ 0.2       & \tb{99.84 $\pm$ 0.1}  & 75.02 $\pm$ 2.6       & 97.05 $\pm$ 0.1       & 0.18\% \\
            \methodsampling $(k=8)$ & 81.32 $\pm$ 1.2       & 72.04 $\pm$ 1.3       & 83.88 $\pm$ 0.7       & 82.05 $\pm$ 0.9       & 90.93 $\pm$ 0.1       & 80.14 $\pm$ 11.2      & 80.12 $\pm$ 0.2       & 99.84 $\pm$ 0.1       & 75.09 $\pm$ 2.5       & 97.05 $\pm$ 0.1       & 0.14\% \\
            \methodsampling $(k=16)$& 81.49 $\pm$ 1.1       & 72.02 $\pm$ 1.2       & 84.23 $\pm$ 0.7       & 82.10 $\pm$ 0.9       & 90.97 $\pm$ 0.1       & 80.10 $\pm$ 11.1      & 80.13 $\pm$ 0.2       & 99.84 $\pm$ 0.1       & 75.16 $\pm$ 2.5       & 97.07 $\pm$ 0.1       & 0.06\% \\
            \method all negatives   & \tb{81.57 $\pm$ 1.1}  & 72.02 $\pm$ 1.2       & \tb{84.26 $\pm$ 0.6}  & \tb{82.15 $\pm$ 0.9}  & \tb{91.12 $\pm$ 0.1}  & \tb{80.25 $\pm$ 11.2} & \tb{80.14 $\pm$ 0.2}  & 99.83 $\pm$ 0.1       & \tb{75.23 $\pm$ 2.4}  & \tb{97.08 $\pm$ 0.1}  & -         \\
            \bottomrule
        \end{tabular}
    }
\end{table*}

Table~\ref{tab:accuracy} summarizes the empirical performance of all methods.
Overall, our proposed method achieves the strongest performance across all datasets.
In most cases, \method outperforms its unsupervised baselines by significant margins, and also outperforms the models trained with label supervision.
Below, we make three notable observations.


First, applying graph contrastive learning methods, such as \Nodetovec, \DGI, and \GRACE, to hypergraph datasets is less effective. 
They show significantly lower accuracy compared to \method.
This is because converting hypergraphs to graphs via clique expansion involves a loss of structural information~\citep{dong2020hnhn}.
Especially, the Zoo dataset has large maximum and average hyperedge sizes (see Appendix~\ref{sec:app:dataset}). 
When clique expansion is performed, a nearly complete graph, where most of the nodes are pairwisely connected to each other, is obtained, and thus most of the structural information is lost, resulting in significant performance degradation.

Second, rather than just using node-level contrast, considering the different types of contrast (i.e., group- and membership-level contrast) together can help improve performance.
We propose and evaluate two model variants, denoted as \methodN and \methodNE, which use only node-level contrast and node- and group-level contrast, respectively, to validate the effect of each type of contrast.
From Table~\ref{tab:accuracy}, we note that the more types of contrast we use, the better the performance tends to be.
To be more specific, we analyze the effectiveness of each type of contrast (i.e., $\calL_{n}$, $\calL_{g}$, and $\calL_{m}$) on the node classification task in Table~\ref{tab:comparison}.
We conduct experiments on all combinations of all types of contrast.
The results show that using all types of contrast achieves the best performance in most cases as they are complementarily reinforcing each other (see Section~\ref{sec:method:proposed} for motivating examples of how different types of contrast can be helpful for node representation learning).
In most cases, using a combination of any two types of contrast is more powerful than using only one.
It is noteworthy that while membership-level contrast causes model collapse\footnote{Model collapse~\citep{zhu2021improving} indicates that the model cannot significantly outperform or even underperform Random-Init. The qualitative analysis of the collapsed models is provided in Appendix~\ref{sec:app:qualitative:collapse}.} (especially for the Citeseer, DBLP, and Mushroom datasets) when used alone, it boosts performance when used with node- or group-level contrast.

Lastly, in Table~\ref{tab:comparison}, we note that group-level contrast is more crucial than node-level contrast for the Citeseer dataset (marked with asterisk), even though the downstream task is node-level.
This result empirically supports our motivations mentioned in Section~\ref{sec:intro}.

To sum up, the superior performance of \method demonstrates that it produces highly generalized representations.
More ablation studies and sensitivity analysis on hyperparameters used in \method are provided in Appendix~\ref{sec:app:exp}.


\paragraph{Robustness to the number of negatives.}
To analyze how the number of negative samples influences the node classification performance, we propose an approximation of \method's objective called \methodsampling.
Here, instead of constructing the contrastive loss with all negatives, we randomly subsample $k$ negatives across the hypergraph for node- and group-level contrast, respectively, at every gradient step.
Our results in Table~\ref{tab:subsampling} show that \method is very robust to the number of negatives; even if only two negative samples are used for node- and group-level contrast, the performance degradation is less than 1\%, still outperforming the best performing unsupervised baseline method, \HHGR, by great margins.
Additionally, the results indicate that the random negative sampling is sufficiently effective for \method, and there is no need to select hard negatives, which incur additional computational costs.

\begin{table*}[!t]
    \setlength{\tabcolsep}{5pt}
    \caption{
    Evaluation of the embeddings learned by unsupervised methods using k-means clustering.
    As a na\"ive baseline method, expressed by `features', we only use the node features as an input of k-means.
    All metrics are normalized by multiplying 100.
    Larger NMI and F1 indicate better performance, and A.R. denotes average ranking.
    \method ranks first in clustering performance.
    }
    \label{tab:clustering}
    \centering
    \scalebox{0.71}{
        \begin{tabular}{lcccccccccccccccccccccc}
            \toprule
            \multirow{2}{*}[-1mm]{Method} & \multicolumn{2}{c}{Cora-C} & \multicolumn{2}{c}{Citeseer} & \multicolumn{2}{c}{Pubmed} & \multicolumn{2}{c}{Cora-A} & \multicolumn{2}{c}{DBLP} & \multicolumn{2}{c}{Zoo} & \multicolumn{2}{c}{20News} & \multicolumn{2}{c}{Mushroom} & \multicolumn{2}{c}{NTU2012} & \multicolumn{2}{c}{ModelNet40} & \hspace{-1mm}\multirow{2}{*}[-1mm]{A.R.$\downarrow$}\hspace{-1mm} \\ 
            \cmidrule(lr){2-3} \cmidrule(lr){4-5} \cmidrule(lr){6-7} \cmidrule(lr){8-9} \cmidrule(lr){10-11} \cmidrule(lr){12-13} \cmidrule(lr){14-15} \cmidrule(lr){16-17} \cmidrule(lr){18-19} \cmidrule(lr){20-21}
             & NMI$\uparrow$ & F1$\uparrow$ & NMI$\uparrow$ & F1$\uparrow$ & NMI$\uparrow$ & F1$\uparrow$ & NMI$\uparrow$ & F1$\uparrow$ & NMI$\uparrow$ & F1$\uparrow$ & NMI$\uparrow$ & F1$\uparrow$ & NMI$\uparrow$ & F1$\uparrow$ & NMI$\uparrow$ & F1$\uparrow$ & NMI$\uparrow$ & F1$\uparrow$ & NMI$\uparrow$ & F1$\uparrow$ \\
            \midrule
            \midrule
            features                & 20.0      & 28.8      & 21.5      & 36.1      & 19.5      & \tb{53.4} & 17.2      & 29.2      & 37.0      & 47.3      & 78.3      & 77.3      & 15.7      & 41.1          & \tb{36.6} & \tb{72.4}     & 81.7      & 69.0      & 90.6      & 86.5      & 3.8       \\ 
            \Nodetovec\!$^{\star}$  & 39.1      & 44.5      & 24.5      & 38.5      & 23.1      & 40.1      & 16.0      & 34.1      & 32.4      & 37.8      & 11.5      & 41.6      & 8.7       & 26.6          & 1.6       & 44.0          & 78.3      & 57.7      & 72.9      & 53.1      & 5.0       \\ 
            \DGI\!$^{\star}$        & \tb{54.8} & \ul{60.1} & 40.1      & 51.7      & \tb{30.4} & 53.0      & 45.2      & \ul{52.5} & 58.0      & 57.7      & 13.0      & 13.8      & \multicolumn{2}{c}{OOM}   & \multicolumn{2}{c}{OOM}   & 79.6      & 61.7      & 85.0      & 73.7      & 3.1       \\ 
            \GRACE\!$^{\star}$      & 44.4      & 45.6      & 33.3      & 45.7      & 16.7      & 41.9      & 37.9      & 43.3      & 16.7      & 41.9      & 7.3       & 29.4      & \multicolumn{2}{c}{OOM}   & \multicolumn{2}{c}{OOM}   & 74.6      & 47.5      & 79.4      & 59.9      & 4.9       \\ 
            \HHGR                   & 51.0      & 56.8      & \ul{41.1} & \ul{53.1} & 27.7      & \ul{53.2} & \ul{45.4} & 52.3      & \ul{60.3} & \ul{62.7} & \ul{90.9} & \tb{91.1} & \tb{39.0} & \tb{58.7}     & \ul{18.6} & 60.6          & \ul{82.7} & \ul{71.2} & \ul{91.0} & \ul{90.6} & \ul{2.1}  \\ 
            \textbf{\method}        & \ul{54.5} & \tb{60.6} & \tb{44.1} & \tb{57.4} & \ul{30.0} & 51.7      & \tb{49.8} & \tb{56.7} & \tb{63.1} & \tb{63.0} & \tb{91.2} & \ul{89.3} & \ul{35.6} & \ul{54.2}     & 3.8       & \ul{65.1}     & \tb{83.2} & \tb{71.5} & \tb{95.7} & \tb{94.7} & \tb{1.6}  \\ 
            \bottomrule
        \end{tabular}
    }
\end{table*}

\begin{table}[!t]
    \caption{
    t-SNE plots of the node representations produced by \method and its two variants.
    The node embeddings of \method exhibits the most distinct clusters with the help of group and membership contrast, as measured numerically by the Silhouette score (the higher, the better).
    }
    \label{tab:tsne}
    \centering
    \scalebox{0.93}{
        \begin{tabular}{ccc}
            \toprule
            & Cora-C & Citeseer \\
            \midrule
            \midrule
            \rotatebox{90}{\hspace{9mm}\methodN} & 
            \hspace{-4mm} \includegraphics[width=0.50\linewidth]{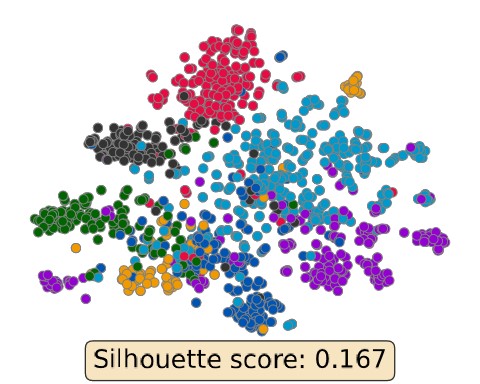} \hspace{-4mm} &
            \hspace{-4mm} \includegraphics[width=0.50\linewidth]{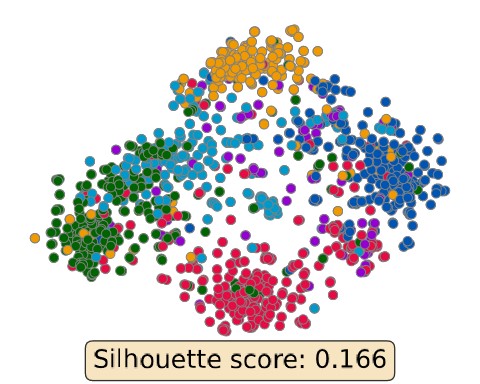} \hspace{-4mm} \\
            \midrule
            \rotatebox{90}{\hspace{8mm}\methodNE} &
            \hspace{-4mm} \includegraphics[width=0.50\linewidth]{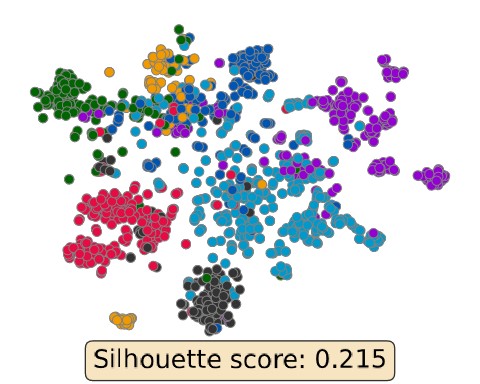} \hspace{-4mm} &
            \hspace{-4mm} \includegraphics[width=0.50\linewidth]{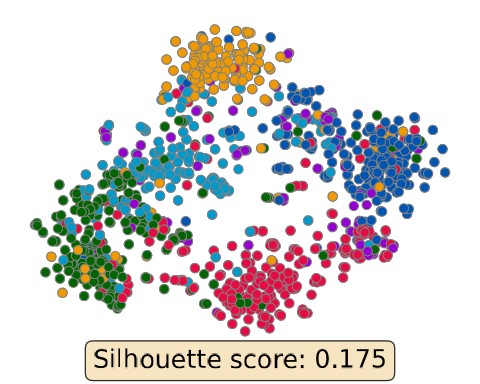} \hspace{-4mm} \\
            \midrule
            \rotatebox{90}{\hspace{11mm}\textbf{\method}} &
            \hspace{-4mm} \includegraphics[width=0.50\linewidth]{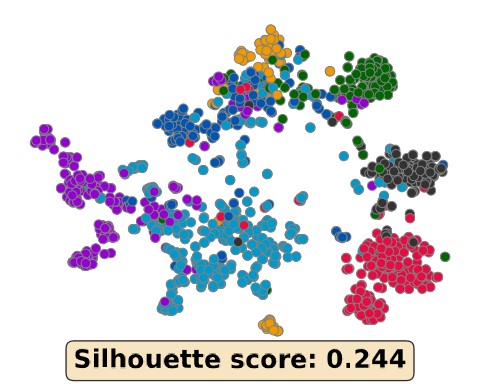} \hspace{-4mm} &
            \hspace{-4mm} \includegraphics[width=0.50\linewidth]{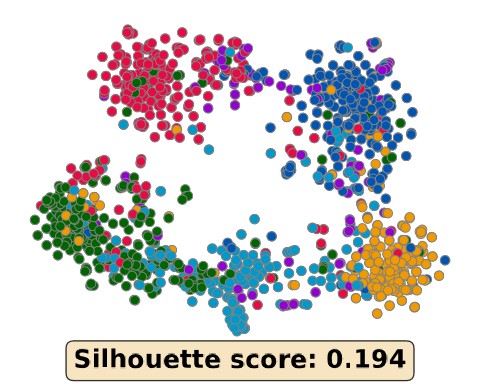} \hspace{-4mm} \\
            \bottomrule
        \end{tabular}
    }
\end{table}

\subsection{Performance on Clustering}
\label{sec:exp:clustering}
To show how well node representations trained with \method generalize across various downstream tasks, we evaluate the representations on the clustering task by k-means as described in Section~\ref{sec:exp:setup}.
We use the node labels as ground truth for the clusters.
To evaluate the clusters generated by k-means, we measure the agreement between the true labels and the cluster assignments by two metrics: Normalized Mutual Information (NMI) and pairwise F1 score.
Table~\ref{tab:clustering} summarizes the empirical performance.
Our results show that \method achieves strong clustering performance in terms of all metrics across all datasets (1st place in terms of the average rank).
This is because the node embeddings learned by \method simultaneously preserve local and community structural information by fully utilizing group-level contrast.



\subsection{Qualitative Analysis}
\label{sec:exp:tsne}

To represent and compare the quality of embeddings intuitively, Table~\ref{tab:tsne} shows t-SNE~\citep{van2008visualizing} plots of the node embeddings produced by \method and its two variants (i.e., \methodN and \methodNE) on the Citeseer and Cora Co-citation dataset.
As expected from the quantitative results, the 2-D projection of embeddings learned by \method shows visually and numerically (based on the Silhouette score~\citep{rousseeuw1987silhouettes}) more distinguishable clusters than those obtained by its two variants.
In Appendix~\ref{sec:app:qualitative}, we give additional qualitative analysis.

\section{Conclusion}
\label{sec:conclusion}


In this paper, we proposed \method, a novel hypergraph contrastive representation learning approach. 
We summarize our contributions as follows:
\begin{itemize}[leftmargin=10pt,itemsep=1pt,topsep=0pt]
    \item We proposed the use of tri-directional contrast, which is a combination of node-, group-, and membership-level contrast, that consistently and substantially improves the quality of the learned embeddings. 
    \item We achieved state-of-the-art results in node classification on hypergraphs by using tri-directional contrast together with our data augmentation schemes. Moreover, we verified the surprising effectiveness of uniform negative sampling for our use cases.
    \item We demonstrated the superiority of \method by conducting extensive experiments using 14 baseline approaches, 10 datasets, and two tasks.
\end{itemize}

\paragraph{Acknowledgements.}
This work was supported by Samsung Electronics Co., Ltd. and Institute of Information \& Communications Technology Planning \& Evaluation (IITP)
grant funded by the Korea government (MSIT) (No. 2022-0-00157,
Robust, Fair, Extensible Data-Centric Continual Learning) (No. 2019-
0-00075, Artificial Intelligence Graduate School Program (KAIST)).

\bibliography{hypergraph.bib}

\begin{thebibliography}{66}
\providecommand{\natexlab}[1]{#1}

\bibitem[{Arthur and Vassilvitskii(2006)}]{arthur2006k}
Arthur, D.; and Vassilvitskii, S. 2006.
\newblock k-means++: The advantages of careful seeding.
\newblock Technical report, Stanford.

\bibitem[{Arya et~al.(2020)Arya, Gupta, Rudinac, and
  Worring}]{arya2020hypersage}
Arya, D.; Gupta, D.~K.; Rudinac, S.; and Worring, M. 2020.
\newblock Hypersage: Generalizing inductive representation learning on
  hypergraphs.
\newblock \emph{arXiv:2010.04558}.

\bibitem[{Bai, Zhang, and Torr(2021)}]{bai2021hypergraph}
Bai, S.; Zhang, F.; and Torr, P.~H. 2021.
\newblock Hypergraph convolution and hypergraph attention.
\newblock \emph{Pattern Recognition}, 110: 107637.

\bibitem[{Benson et~al.(2018)Benson, Abebe, Schaub, Jadbabaie, and
  Kleinberg}]{benson2018simplicial}
Benson, A.~R.; Abebe, R.; Schaub, M.~T.; Jadbabaie, A.; and Kleinberg, J. 2018.
\newblock Simplicial closure and higher-order link prediction.
\newblock \emph{PNAS}, 115(48).

\bibitem[{Benson, Gleich, and Leskovec(2016)}]{benson2016higher}
Benson, A.~R.; Gleich, D.~F.; and Leskovec, J. 2016.
\newblock Higher-order organization of complex networks.
\newblock \emph{Science}, 353(6295): 163--166.

\bibitem[{Chen et~al.(2003)Chen, Tian, Shen, and Ouhyoung}]{chen2003visual}
Chen, D.-Y.; Tian, X.-P.; Shen, Y.-T.; and Ouhyoung, M. 2003.
\newblock On visual similarity based 3D model retrieval.
\newblock \emph{Computer graphics forum}, 22(3): 223--232.

\bibitem[{Chen et~al.(2020)Chen, Kornblith, Norouzi, and
  Hinton}]{chen2020simple}
Chen, T.; Kornblith, S.; Norouzi, M.; and Hinton, G. 2020.
\newblock A simple framework for contrastive learning of visual
  representations.
\newblock In \emph{ICML}, 1597--1607.

\bibitem[{Chien et~al.(2022)Chien, Pan, Peng, and Milenkovic}]{chien2021you}
Chien, E.; Pan, C.; Peng, J.; and Milenkovic, O. 2022.
\newblock You are AllSet: A Multiset Function Framework for Hypergraph Neural
  Networks.
\newblock In \emph{ICLR}.

\bibitem[{Clevert, Unterthiner, and Hochreiter(2016)}]{clevert2015fast}
Clevert, D.-A.; Unterthiner, T.; and Hochreiter, S. 2016.
\newblock Fast and accurate deep network learning by exponential linear units
  (elus).
\newblock In \emph{ICLR}.

\bibitem[{Do et~al.(2020)Do, Yoon, Hooi, and Shin}]{do2020structural}
Do, M.~T.; Yoon, S.-e.; Hooi, B.; and Shin, K. 2020.
\newblock Structural patterns and generative models of real-world hypergraphs.
\newblock In \emph{KDD}, 176--186.

\bibitem[{Dong, Sawin, and Bengio(2020)}]{dong2020hnhn}
Dong, Y.; Sawin, W.; and Bengio, Y. 2020.
\newblock HNHN: Hypergraph networks with hyperedge neurons.
\newblock In \emph{ICML Workshop on Graph Representation Learning and Beyond}.

\bibitem[{Dua and Graff(2017)}]{Dua:2019}
Dua, D.; and Graff, C. 2017.
\newblock {UCI} Machine Learning Repository.

\bibitem[{Feng et~al.(2019)Feng, You, Zhang, Ji, and Gao}]{feng2019hypergraph}
Feng, Y.; You, H.; Zhang, Z.; Ji, R.; and Gao, Y. 2019.
\newblock Hypergraph neural networks.
\newblock In \emph{AAAI}, volume~33, 3558--3565.

\bibitem[{Feng et~al.(2018)Feng, Zhang, Zhao, Ji, and Gao}]{feng2018gvcnn}
Feng, Y.; Zhang, Z.; Zhao, X.; Ji, R.; and Gao, Y. 2018.
\newblock Gvcnn: Group-view convolutional neural networks for 3d shape
  recognition.
\newblock In \emph{CVPR}, 264--272.

\bibitem[{Fey and Lenssen(2019)}]{fey2019fast}
Fey, M.; and Lenssen, J.~E. 2019.
\newblock Fast graph representation learning with PyTorch Geometric.
\newblock \emph{arXiv:1903.02428}.

\bibitem[{Gao, Yao, and Chen(2021)}]{gao2021simcse}
Gao, T.; Yao, X.; and Chen, D. 2021.
\newblock Simcse: Simple contrastive learning of sentence embeddings.
\newblock In \emph{EMNLP}.

\bibitem[{Glorot and Bengio(2010)}]{glorot2010understanding}
Glorot, X.; and Bengio, Y. 2010.
\newblock Understanding the difficulty of training deep feedforward neural
  networks.
\newblock In \emph{AISTATS}, 249--256.

\bibitem[{Grover and Leskovec(2016)}]{grover2016node2vec}
Grover, A.; and Leskovec, J. 2016.
\newblock node2vec: Scalable feature learning for networks.
\newblock In \emph{KDD}, 855--864.

\bibitem[{Hassani and Khasahmadi(2020)}]{hassani2020contrastive}
Hassani, K.; and Khasahmadi, A.~H. 2020.
\newblock Contrastive multi-view representation learning on graphs.
\newblock In \emph{ICML}, 4116--4126.

\bibitem[{He et~al.(2020)He, Fan, Wu, Xie, and Girshick}]{he2020momentum}
He, K.; Fan, H.; Wu, Y.; Xie, S.; and Girshick, R. 2020.
\newblock Momentum contrast for unsupervised visual representation learning.
\newblock In \emph{CVPR}, 9729--9738.

\bibitem[{He et~al.(2015)He, Zhang, Ren, and Sun}]{he2015delving}
He, K.; Zhang, X.; Ren, S.; and Sun, J. 2015.
\newblock Delving deep into rectifiers: Surpassing human-level performance on
  imagenet classification.
\newblock In \emph{ICCV}, 1026--1034.

\bibitem[{Hein et~al.(2013)Hein, Setzer, Jost, and Rangapuram}]{hein2013total}
Hein, M.; Setzer, S.; Jost, L.; and Rangapuram, S.~S. 2013.
\newblock The total variation on hypergraphs-learning on hypergraphs revisited.
\newblock In \emph{NIPS}.

\bibitem[{Hjelm et~al.(2019)Hjelm, Fedorov, Lavoie-Marchildon, Grewal, Bachman,
  Trischler, and Bengio}]{hjelm2018learning}
Hjelm, R.~D.; Fedorov, A.; Lavoie-Marchildon, S.; Grewal, K.; Bachman, P.;
  Trischler, A.; and Bengio, Y. 2019.
\newblock Learning deep representations by mutual information estimation and
  maximization.
\newblock In \emph{ICLR}.

\bibitem[{Huang and Yang(2021)}]{huang2021unignn}
Huang, J.; and Yang, J. 2021.
\newblock Unignn: a unified framework for graph and hypergraph neural networks.
\newblock In \emph{IJCAI}.

\bibitem[{Jaiswal et~al.(2020)Jaiswal, Babu, Zadeh, Banerjee, and
  Makedon}]{jaiswal2020survey}
Jaiswal, A.; Babu, A.~R.; Zadeh, M.~Z.; Banerjee, D.; and Makedon, F. 2020.
\newblock A survey on contrastive self-supervised learning.
\newblock \emph{Technologies}, 9(1): 2.

\bibitem[{Kingma and Ba(2015)}]{kingma2014adam}
Kingma, D.~P.; and Ba, J. 2015.
\newblock Adam: A method for stochastic optimization.
\newblock In \emph{ICLR}.

\bibitem[{Kipf and Welling(2017)}]{kipf2016semi}
Kipf, T.~N.; and Welling, M. 2017.
\newblock Semi-supervised classification with graph convolutional networks.
\newblock In \emph{ICLR}.

\bibitem[{Lee, Choe, and Shin(2022)}]{lee2022hashnwalk}
Lee, G.; Choe, M.; and Shin, K. 2022.
\newblock HashNWalk: Hash and Random Walk Based Anomaly Detection in Hyperedge
  Streams.
\newblock In \emph{IJCAI}.

\bibitem[{Lee, Ko, and Shin(2020)}]{lee2020hypergraph}
Lee, G.; Ko, J.; and Shin, K. 2020.
\newblock Hypergraph motifs: concepts, algorithms, and discoveries.
\newblock In \emph{Proceedings of the VLDB Endowment}.

\bibitem[{Lee et~al.(2019)Lee, Lee, Kim, Kosiorek, Choi, and Teh}]{lee2019set}
Lee, J.; Lee, Y.; Kim, J.; Kosiorek, A.; Choi, S.; and Teh, Y.~W. 2019.
\newblock Set transformer: A framework for attention-based
  permutation-invariant neural networks.
\newblock In \emph{ICML}, 3744--3753.

\bibitem[{Li, Jing, and Tong(2022)}]{li2022graph}
Li, B.; Jing, B.; and Tong, H. 2022.
\newblock Graph Communal Contrastive Learning.
\newblock In \emph{WWW}, 1203--1213.

\bibitem[{Li and Milenkovic(2018)}]{li2018submodular}
Li, P.; and Milenkovic, O. 2018.
\newblock Submodular hypergraphs: p-laplacians, cheeger inequalities and
  spectral clustering.
\newblock In \emph{ICML}, 3014--3023.

\bibitem[{Liu et~al.(2021)Liu, Zhang, Hou, Mian, Wang, Zhang, and
  Tang}]{liu2021self}
Liu, X.; Zhang, F.; Hou, Z.; Mian, L.; Wang, Z.; Zhang, J.; and Tang, J. 2021.
\newblock Self-supervised learning: Generative or contrastive.
\newblock \emph{TKDE}.

\bibitem[{Liu et~al.(2022)Liu, Jin, Pan, Zhou, Zheng, Xia, and
  Yu}]{liu2022graph}
Liu, Y.; Jin, M.; Pan, S.; Zhou, C.; Zheng, Y.; Xia, F.; and Yu, P. 2022.
\newblock Graph self-supervised learning: A survey.
\newblock \emph{TKDE}.

\bibitem[{Lloyd(1982)}]{lloyd1982least}
Lloyd, S. 1982.
\newblock Least squares quantization in PCM.
\newblock \emph{IEEE Transactions on Information Theory}, 28(2): 129--137.

\bibitem[{Loshchilov and Hutter(2019)}]{loshchilov2017decoupled}
Loshchilov, I.; and Hutter, F. 2019.
\newblock Decoupled weight decay regularization.
\newblock In \emph{ICLR}.

\bibitem[{Oord, Li, and Vinyals(2018)}]{oord2018representation}
Oord, A. v.~d.; Li, Y.; and Vinyals, O. 2018.
\newblock Representation learning with contrastive predictive coding.
\newblock \emph{arXiv:1807.03748}.

\bibitem[{Paszke et~al.(2019)Paszke, Gross, Massa, Lerer, Bradbury, Chanan,
  Killeen, Lin, Gimelshein, Antiga et~al.}]{paszke2019pytorch}
Paszke, A.; Gross, S.; Massa, F.; Lerer, A.; Bradbury, J.; Chanan, G.; Killeen,
  T.; Lin, Z.; Gimelshein, N.; Antiga, L.; et~al. 2019.
\newblock Pytorch: An imperative style, high-performance deep learning library.
\newblock In \emph{NeurIPS}.

\bibitem[{Peng et~al.(2020)Peng, Huang, Luo, Zheng, Rong, Xu, and
  Huang}]{peng2020graph}
Peng, Z.; Huang, W.; Luo, M.; Zheng, Q.; Rong, Y.; Xu, T.; and Huang, J. 2020.
\newblock Graph representation learning via graphical mutual information
  maximization.
\newblock In \emph{WWW}, 259--270.

\bibitem[{Qiu et~al.(2020)Qiu, Chen, Dong, Zhang, Yang, Ding, Wang, and
  Tang}]{qiu2020gcc}
Qiu, J.; Chen, Q.; Dong, Y.; Zhang, J.; Yang, H.; Ding, M.; Wang, K.; and Tang,
  J. 2020.
\newblock Gcc: Graph contrastive coding for graph neural network pre-training.
\newblock In \emph{KDD}, 1150--1160.

\bibitem[{Rong et~al.(2020)Rong, Huang, Xu, and Huang}]{rong2019dropedge}
Rong, Y.; Huang, W.; Xu, T.; and Huang, J. 2020.
\newblock Dropedge: Towards deep graph convolutional networks on node
  classification.
\newblock In \emph{ICLR}.

\bibitem[{Rossi and Ahmed(2015)}]{rossi2015network}
Rossi, R.; and Ahmed, N. 2015.
\newblock The network data repository with interactive graph analytics and
  visualization.
\newblock In \emph{AAAI}.

\bibitem[{Rousseeuw(1987)}]{rousseeuw1987silhouettes}
Rousseeuw, P.~J. 1987.
\newblock Silhouettes: a graphical aid to the interpretation and validation of
  cluster analysis.
\newblock \emph{Journal of Computational and Applied Mathematics}, 20: 53--65.

\bibitem[{Sen et~al.(2008)Sen, Namata, Bilgic, Getoor, Galligher, and
  Eliassi-Rad}]{sen2008collective}
Sen, P.; Namata, G.; Bilgic, M.; Getoor, L.; Galligher, B.; and Eliassi-Rad, T.
  2008.
\newblock Collective classification in network data.
\newblock \emph{AI magazine}, 29(3): 93--93.

\bibitem[{Su et~al.(2015)Su, Maji, Kalogerakis, and
  Learned-Miller}]{su2015multi}
Su, H.; Maji, S.; Kalogerakis, E.; and Learned-Miller, E. 2015.
\newblock Multi-view convolutional neural networks for 3d shape recognition.
\newblock In \emph{ICCV}, 945--953.

\bibitem[{Thakoor et~al.(2022)Thakoor, Tallec, Azar, Azabou, Dyer, Munos,
  Veli{\v{c}}kovi{\'c}, and Valko}]{thakoor2021large}
Thakoor, S.; Tallec, C.; Azar, M.~G.; Azabou, M.; Dyer, E.~L.; Munos, R.;
  Veli{\v{c}}kovi{\'c}, P.; and Valko, M. 2022.
\newblock Large-scale representation learning on graphs via bootstrapping.
\newblock In \emph{ICLR}.

\bibitem[{Tschannen et~al.(2019)Tschannen, Djolonga, Rubenstein, Gelly, and
  Lucic}]{tschannen2019mutual}
Tschannen, M.; Djolonga, J.; Rubenstein, P.~K.; Gelly, S.; and Lucic, M. 2019.
\newblock On Mutual Information Maximization for Representation Learning.
\newblock In \emph{ICLR}.

\bibitem[{Van~der Maaten and Hinton(2008)}]{van2008visualizing}
Van~der Maaten, L.; and Hinton, G. 2008.
\newblock Visualizing data using t-SNE.
\newblock \emph{Journal of Machine Learning Research}, 9(11).

\bibitem[{Veldt, Benson, and Kleinberg(2021)}]{veldt2021higher}
Veldt, N.; Benson, A.~R.; and Kleinberg, J. 2021.
\newblock Higher-order homophily is combinatorially impossible.
\newblock \emph{arXiv:2103.11818}.

\bibitem[{Veli{\v{c}}kovi{\'c} et~al.(2018{\natexlab{a}})Veli{\v{c}}kovi{\'c},
  Cucurull, Casanova, Romero, Li{\`o}, and Bengio}]{velivckovic2018graph}
Veli{\v{c}}kovi{\'c}, P.; Cucurull, G.; Casanova, A.; Romero, A.; Li{\`o}, P.;
  and Bengio, Y. 2018{\natexlab{a}}.
\newblock Graph Attention Networks.
\newblock In \emph{ICLR}.

\bibitem[{Veli{\v{c}}kovi{\'c} et~al.(2018{\natexlab{b}})Veli{\v{c}}kovi{\'c},
  Fedus, Hamilton, Li{\`o}, Bengio, and Hjelm}]{velivckovic2018deep}
Veli{\v{c}}kovi{\'c}, P.; Fedus, W.; Hamilton, W.~L.; Li{\`o}, P.; Bengio, Y.;
  and Hjelm, R.~D. 2018{\natexlab{b}}.
\newblock Deep Graph Infomax.
\newblock In \emph{ICLR}.

\bibitem[{Wang and Liu(2021)}]{wang2021understanding}
Wang, F.; and Liu, H. 2021.
\newblock Understanding the behaviour of contrastive loss.
\newblock In \emph{CVPR}, 2495--2504.

\bibitem[{Wu et~al.(2015)Wu, Song, Khosla, Yu, Zhang, Tang, and
  Xiao}]{wu20153d}
Wu, Z.; Song, S.; Khosla, A.; Yu, F.; Zhang, L.; Tang, X.; and Xiao, J. 2015.
\newblock 3d shapenets: A deep representation for volumetric shapes.
\newblock In \emph{CVPR}, 1912--1920.

\bibitem[{Xia et~al.(2021)Xia, Yin, Yu, Wang, Cui, and Zhang}]{xia2021self}
Xia, X.; Yin, H.; Yu, J.; Wang, Q.; Cui, L.; and Zhang, X. 2021.
\newblock Self-supervised hypergraph convolutional networks for session-based
  recommendation.
\newblock In \emph{AAAI}, volume~35, 4503--4511.

\bibitem[{Yadati et~al.(2019)Yadati, Nimishakavi, Yadav, Nitin, Louis, and
  Talukdar}]{yadati2019hypergcn}
Yadati, N.; Nimishakavi, M.; Yadav, P.; Nitin, V.; Louis, A.; and Talukdar, P.
  2019.
\newblock Hypergcn: A new method for training graph convolutional networks on
  hypergraphs.
\newblock In \emph{NeurIPS}.

\bibitem[{Yang et~al.(2019)Yang, Qu, Yang, and
  Cudre-Mauroux}]{yang2019revisiting}
Yang, D.; Qu, B.; Yang, J.; and Cudre-Mauroux, P. 2019.
\newblock Revisiting user mobility and social relationships in lbsns: a
  hypergraph embedding approach.
\newblock In \emph{WWW}, 2147--2157.

\bibitem[{You et~al.(2020)You, Chen, Sui, Chen, Wang, and Shen}]{you2020graph}
You, Y.; Chen, T.; Sui, Y.; Chen, T.; Wang, Z.; and Shen, Y. 2020.
\newblock Graph contrastive learning with augmentations.
\newblock In \emph{NeurIPS}.

\bibitem[{Yu et~al.(2021)Yu, Yin, Li, Wang, Hung, and Zhang}]{yu2021self}
Yu, J.; Yin, H.; Li, J.; Wang, Q.; Hung, N. Q.~V.; and Zhang, X. 2021.
\newblock Self-supervised multi-channel hypergraph convolutional network for
  social recommendation.
\newblock In \emph{WWW}, 413--424.

\bibitem[{Zaheer et~al.(2017)Zaheer, Kottur, Ravanbakhsh, Poczos,
  Salakhutdinov, and Smola}]{zaheer2017deep}
Zaheer, M.; Kottur, S.; Ravanbakhsh, S.; Poczos, B.; Salakhutdinov, R.~R.; and
  Smola, A.~J. 2017.
\newblock Deep sets.
\newblock In \emph{NIPS}.

\bibitem[{Zhang et~al.(2021)Zhang, Gao, Yu, Guo, Li, and Yin}]{zhang2021double}
Zhang, J.; Gao, M.; Yu, J.; Guo, L.; Li, J.; and Yin, H. 2021.
\newblock Double-Scale Self-Supervised Hypergraph Learning for Group
  Recommendation.
\newblock In \emph{CIKM}, 2557--2567.

\bibitem[{Zhang et~al.(2022)Zhang, Li, Xiao, Xu, Rong, Huang, and
  Bian}]{zhang2022hypergraph}
Zhang, J.; Li, F.; Xiao, X.; Xu, T.; Rong, Y.; Huang, J.; and Bian, Y. 2022.
\newblock Hypergraph Convolutional Networks via Equivalency between Hypergraphs
  and Undirected Graphs.
\newblock \emph{arXiv:2203.16939}.

\bibitem[{Zheng et~al.(2019)Zheng, Zhu, Tang, and Wang}]{zheng2019gene}
Zheng, X.; Zhu, W.; Tang, C.; and Wang, M. 2019.
\newblock Gene selection for microarray data classification via adaptive
  hypergraph embedded dictionary learning.
\newblock \emph{Gene}, 706: 188--200.

\bibitem[{Zhou, Huang, and Sch{\"o}lkopf(2006)}]{zhou2006learning}
Zhou, D.; Huang, J.; and Sch{\"o}lkopf, B. 2006.
\newblock Learning with hypergraphs: Clustering, classification, and embedding.
\newblock In \emph{NIPS}.

\bibitem[{Zhu et~al.(2021{\natexlab{a}})Zhu, Zhao, Liu, Sun, and
  Chen}]{zhu2021improving}
Zhu, R.; Zhao, B.; Liu, J.; Sun, Z.; and Chen, C.~W. 2021{\natexlab{a}}.
\newblock Improving contrastive learning by visualizing feature transformation.
\newblock In \emph{ICCV}, 10306--10315.

\bibitem[{Zhu et~al.(2020)Zhu, Xu, Yu, Liu, Wu, and Wang}]{zhu2020deep}
Zhu, Y.; Xu, Y.; Yu, F.; Liu, Q.; Wu, S.; and Wang, L. 2020.
\newblock Deep graph contrastive representation learning.
\newblock In \emph{ICML Workshop on Graph Representation Learning and Beyond}.

\bibitem[{Zhu et~al.(2021{\natexlab{b}})Zhu, Xu, Yu, Liu, Wu, and
  Wang}]{zhu2021graph}
Zhu, Y.; Xu, Y.; Yu, F.; Liu, Q.; Wu, S.; and Wang, L. 2021{\natexlab{b}}.
\newblock Graph contrastive learning with adaptive augmentation.
\newblock In \emph{WWW}, 2069--2080.

\end{thebibliography}

\clearpage
\appendix

\section{Dataset Details}
\label{sec:app:dataset}
We use 10 benchmark datasets from the existing hypergraph neural networks literature;
these datasets are categorized into
(1) co-citation datasets (Cora, Citeseer, and Pubmed) \footnote{https://linqs.soe.ucsc.edu/data}~\citep{sen2008collective},
(2) co-authorship datasets (Cora \footnote{https://people.cs.umass.edu/~mccallum/data.html} and DBLP \footnote{https://aminer.org/lab-datasets/citation/DBLP-citation-Jan8.tar.bz}~\citep{rossi2015network}), 
(3) computer vision and graphics datasets (NTU2012~\citep{chen2003visual} and ModelNet40~\citep{wu20153d}), and
(4) datasets from the UCI Categorical Machine Learning Repository~\citep{Dua:2019} (Zoo, 20Newsgroups, and Mushroom).
Some basic statistics of the datasets are provided in Table~\ref{tab:dataset}.

The co-citation datasets are composed of a set of papers and their citation links.
To represent a co-citation relationship as a hypergraph, papers become nodes and citation links become hyperedges.
To be specific, the nodes $v_{1},\dots,v_{k}$ compose a hyperedge $e$ when the papers corresponding to $v_{1},\dots,v_{k}$ are referred by the document $e$.
The co-authorship datasets are composed of a set of papers with their authors.
In hypergraphs that model the co-authorship datasets, nodes and hyperedges represent papers and authors, respectively.
Precisely, the nodes $v_{1},\dots,v_{k}$ compose a hyperedge $e$ when the papers corresponding to $v_{1},\dots,v_{k}$ are written by the author $e$.
Features of each node are represented by bag-of-words features from its abstract.
Nodes are labeled with their categories.
The hypergraphs preprocessed from all the co-citation and co-authorship datasets are publicly available with the official implementation of HyperGCN \footnote{https://github.com/malllabiisc/HyperGCN}~\citep{yadati2019hypergcn}.

For visual datasets, the hypergraph construction follows the setting described in ~\citet{feng2019hypergraph}, and the node features are extracted by Group-View Convolutional Neural Network (GVCNN)~\citep{feng2018gvcnn} and Multi-View Convolutional Neural Network (MVCNN)~\citep{su2015multi}. 

In the 20Newsgroups dataset, the TF-IDF representations of news messages are used as the node features. 
In the Mushroom dataset, the node features indicate categorical descriptions of 23 species of mushrooms. 
In the Zoo dataset, the node features are a mix of categorical and numerical measurements describing different animals.

We remove nodes that are not included in any hyperedge (i.e. isolated nodes) from the hypergraphs, because such nodes cause trivial structures in hypergraphs and their predictions would only depend on the features of that node.
For all the datasets, we randomly select 10\%, 10\%, and 80\% of nodes disjointly for the training, validation, and test sets, respectively.
The datasets and train-valid-test splits used in our experiments are provided as supplementary materials.

\begin{table*}[!t]
    \caption{Statistics of datasets used in our experiments.}
    \label{tab:dataset}
    \centering
    \scalebox{0.8}{
        \begin{tabular}{lcccccccccc}
            \toprule
             & Cora-C & Citeseer & Pubmed & Cora-A & DBLP & Zoo & 20News & Mushroom & NTU2012 & ModelNet40 \\
            \midrule
            \midrule
                $\#$ Nodes             & 1,434  & 1,458     & 3,840     & 2,388     & 41,302  & 101     & 16,242    & 8,124     & 2,012     & 12,311  \\
                $\#$ Hyperedges        & 1,579  & 1,079     & 7,963     & 1,072     & 22,363  & 43      & 100       & 298       & 2,012     & 12,311  \\
                $\#$ Memberships       & 4,786  & 3,453     & 34,629    & 4,585     & 99,561  & 1,717   & 65,451    & 40,620    & 10,060    & 61,555  \\
                Avg. hyperedge size    & 3.03   & 3.20      & 4.35      & 4.28      & 4.45    & 39.93   & 654.51    & 136.31    & 5         & 5    \\
                Avg. node degree       & 3.34   & 2.37      & 9.02      & 1.92      & 2.41    & 17.00   & 4.03      & 5.00      & 5         & 5    \\
                Max. hyperedge size    & 5      & 26        & 171       & 43        & 202     & 93      & 2241      & 1808      & 5         & 5     \\
                Max. node degree       & 145    & 88        & 99        & 23        & 18      & 17      & 44        & 5         & 19        & 30      \\
                $\#$ Features          & 1,433  & 3,703     & 500       & 1,433     & 1,425   & 16      & 100       & 22        & 100       & 100   \\
                $\#$ Classes           & 7      & 6         & 3         & 7         & 6       & 7       & 4         & 2         & 67        & 40       \\
            \bottomrule
        \end{tabular}
    }
\end{table*}
\section{Baseline Details}
\label{sec:app:baseline}

We compare our proposed method with various representative baseline approaches that can be categorized into
(1) supervised learning methods (\GCN~\citep{kipf2016semi} and \GAT~\citep{velivckovic2018graph} applied to graphs and \HGNN~\citep{feng2019hypergraph}, \HyperConv~\citep{bai2021hypergraph}, \HNHN~\citep{dong2020hnhn}, \HyperGCN~\citep{yadati2019hypergcn}, \HyperSAGE~\citep{arya2020hypersage}, \UniGCN~\citep{huang2021unignn}, and \AllSetTransformer~\citep{chien2021you} applied directly to hypergraphs), and
(2) unsupervised learning methods (\Nodetovec~\citep{grover2016node2vec}, \DGI~\citep{velivckovic2018deep}, and \GRACE~\citep{zhu2020deep}, which are representative graph contrastive learning methods and \HHGR~\citep{zhang2021double}, which is a hypergraphs contrastive learning method). 
To measure the quality of the inductive biases inherent in the encoder model, we also consider \RandomInit~\citep{velivckovic2018deep,thakoor2021large}, an encoder with the same architecture as \method but with randomly initialized parameters, as a baseline.
Since the methods working on graphs can not be directly applied to hypergraphs, we use them after transforming hypergraphs to graphs via clique expansion.
In the case of \HHGR, it is originally designed for group recommendations with supervisory signals, and therefore it is not directly applicable to node classification tasks.
Thus we slightly modified the algorithm so that it uses only its self-supervised loss.
For all the baseline approaches, we report their performance using their official implementations.
\section{Implementation Details}
\label{sec:app:impl}

\subsection{Infrastructures and Implementations}
\label{sec:app:impl:infra}
All experiments are performed on a server with NVIDIA RTX 3090 Ti GPUs (24GB memory), 256GB of RAM, and two Intel Xeon Silver 4210R Processors.
Our models are implemented using PyTorch 1.11.0~\citep{paszke2019pytorch} and PyTorch Geometric 2.0.4~\citep{fey2019fast}.

\begin{table*}[!t]
    \caption{Hyperparameter settings on each dataset.}
    \label{tab:hyperparameter}
    \centering
    \scalebox{0.80}{
        \begin{tabular}{lcccccccccccc}
            \toprule
            Dataset & $p_{f}$ & $p_{m}$ & $\tau_{n}$ & $\tau_{g}$ & $\tau_{m}$ & $\weightg$ & $\weightm$ & \makecell{Learning\\Rate} & \makecell{Training\\Epochs} & \makecell{Node\\Emb. Size} & \makecell{Hyperedge\\Emb. Size} & \makecell{Projection\\Hidden Size} \\
            \midrule
            \midrule
            Cora-C      & 0.4 & 0.4 & 0.5 & 0.5 & 1.0 & $2^{2}$ & 1 & 5e-4 & 300 & 512 & 512 & 512 \\
            Citeseer    & 0.4 & 0.4 & 1.0 & 1.0 & 0.8 & $2^{2}$ & $2^{1}$ & 5e-5 & 500 & 512 & 512 & 512 \\
            Pubmed      & 0.1 & 0.4 & 0.3 & 0.2 & 0.6 & $2^{2}$ & $2^{1}$ & 5e-4 & 1,000 & 512 & 512 & 512 \\
            Cora-A      & 0.3 & 0.2 & 0.6 & 0.5 & 0.6 & $2^{-1}$ & $2^{-1}$ & 1e-4 & 800 & 512 & 512 & 512 \\
            DBLP        & 0.2 & 0.2 & 0.8 & 0.2 & 1.0 & $2^{-4}$ & $2^{-2}$ & 5e-3 & 500 & 256 & 256 & 256 \\
            
            Zoo         & 0.4 & 0.2 & 0.9 & 0.9 & 1.0 & $2^{1}$ & $2^{1}$ & 1e-3 & 100 & 128 & 128 & 128 \\
            20News      & 0.1 & 0.4 & 0.7 & 0.1 & 1.0 & $2^{-4}$ & $2^{-4}$ & 1e-3 & 500 & 256 & 256 & 256 \\
            Mushroom    & 0.0 & 0.4 & 1.0 & 0.9 & 0.1 & $2^{2}$ & 1 & 1e-3 & 500 & 512 & 512 & 512 \\
            NTU2012     & 0.0 & 0.4 & 1.0 & 0.7 & 0.5 & $2^{-1}$ & $2^{-4}$ & 1e-3 & 200 & 512 & 512 & 512 \\
            ModelNet40  & 0.0 & 0.4 & 0.9 & 0.3 & 0.9 & $2^{-2}$ & $2^{-3}$ & 1e-3 & 200 & 256 & 256 & 256 \\
            \bottomrule
        \end{tabular}
    }
\end{table*}

\subsection{Hyperparameters}
\label{sec:app:impl:param}
As described in Section~\ref{sec:exp:setup}, we use a one-layer mean pooling hypergraph encoder as in Eq.~(\ref{eq:encoder:matrix}) and PReLU~\citep{he2015delving} activation in all the experiments.
Note that, to each node, we add a self-loop which is a hyperedge which contains exactly one node, before the hypergraph is fed into the encoder.
In Appendix~\ref{sec:app:exp:self}, we show that adding self-loops helps to improve the quality of representations.
When constructing the proposed tri-directional contrastive loss, self-loops and empty-hyperedges (i.e., hyperedges with degree zero) are ignored.

In all our experiments, all models are initialized with Glorot initialization~\citep{glorot2010understanding} and trained using the AdamW optimizer~\citep{kingma2014adam,loshchilov2017decoupled} with weight decay set to $10^{-5}$.
We train the model for a fixed number of epochs at which the performance of node classification sufficiently converges.

The augmentation hyperparameters $p_{f}$ and $p_{m}$, which control the sampling process for node feature and membership masking, respectively, are chosen between 0.0 and 0.4 so that the original hypergraph is not overly corrupted.
Some prior works~\citep{zhu2020deep,zhu2021graph} have demonstrated that using a different degree of augmentation for each view shows better results, and we can also adopt different hyperparameters for each augmented view (as mentioned in Section~\ref{sec:method:proposed}).
However, our contributions are orthogonal to this problem, thus we choose the same hyperparameters for two augmented views (i.e., $p_{f,1}=p_{f,2}=p_{f}$ and $p_{m,1}=p_{m,2}=p_{m}$) for simplicity.
In Appendix~\ref{sec:app:augmentation}, we demonstrate that using node feature masking and membership masking together is a reasonable choice.

The three temperature hyperparameters $\tau_{n}$, $\tau_{g}$, and $\tau_{m}$, which control the uniformity of the embedding distribution~\citep{wang2021understanding}, are selected from 0.1 to 1.0, respectively.
The weights $\weightg$ and $\weightm$ are chosen from $[2^{-4},2^{-3},\dots,2^{4}]$, respectively.
The size of node embeddings, hyperedge embeddings, and a hidden layer of projection heads are set to the same values for simplicity.
In Table~\ref{tab:hyperparameter}, we provide hyperparameters we found through a small grid search based on the validation accuracy, as many self-supervised learning methods do~\citep{chen2020simple,zhu2020deep,zhu2021graph,thakoor2021large}.

\section{Hypergraph Augmentations}
\label{sec:app:augmentation}

Generating augmented views is crucial for contrastive learning methods.
Different views provide different contexts or semantics for datasets.
While creating semantically meaningful augmentations is critical for contrastive learning, in the hypergraph domain, it is an underexplored problem than in other domains such as vision.
In the graph domain, simple and effective graph augmentation methods have been proposed, and these are commonly used in graph contrastive learning~\citep{you2020graph,zhu2020deep}.
Borrowing these approaches, in this section, we analyze four types of augmentation (i.e., node masking, hyperedge masking, membership masking, and node feature masking), which are naturally applicable to hypergraphs, along with \method.
\begin{itemize}[leftmargin=2mm]
    \item \textbf{Node masking}: randomly mask a portion of nodes in the original hypergraph. Formally, we use a binary mask of size $|V|$ where each element is sampled from a Bernoulli distribution $\calB(1-p_{n})$ to mask nodes.
    \item \textbf{Hyperedge masking}: randomly mask a portion of hyperedges in the original hypergraph. Precisely, we use a binary mask of size $|E|$ where each element is sampled from a Bernoulli distribution $\calB(1-p_{e})$ to mask hyperedges.
    \item \textbf{Membership masking}: randomly mask a portion of node-hyperedge memberships in the original hypergraph. In particular, we use a binary mask of size $K=nnz(\matH)$ where each element is sampled from a Bernoulli distribution $\calB(1-p_{m})$ to mask node-hyperedge memberships.
    \item \textbf{Node feature masking}: randomly mask a portion of dimensions with zeros in node features. Specifically, we generate a single random binary mask of size $F$ where each entry is sampled from a Bernoulli distribution $\calB(1-p_{f})$, and use it to mask features of all nodes in the hypergraph.
\end{itemize}
The degree of augmentation can be controlled by $p_{n}$, $p_{e}$, $p_{m}$, and $p_{f}$.
These masking methods corrupt the hypergraph structure, except for node feature masking, which impairs the hypergraph attributes.

\begin{figure}[!t]
    \centering
    \includegraphics[width=0.8\linewidth]{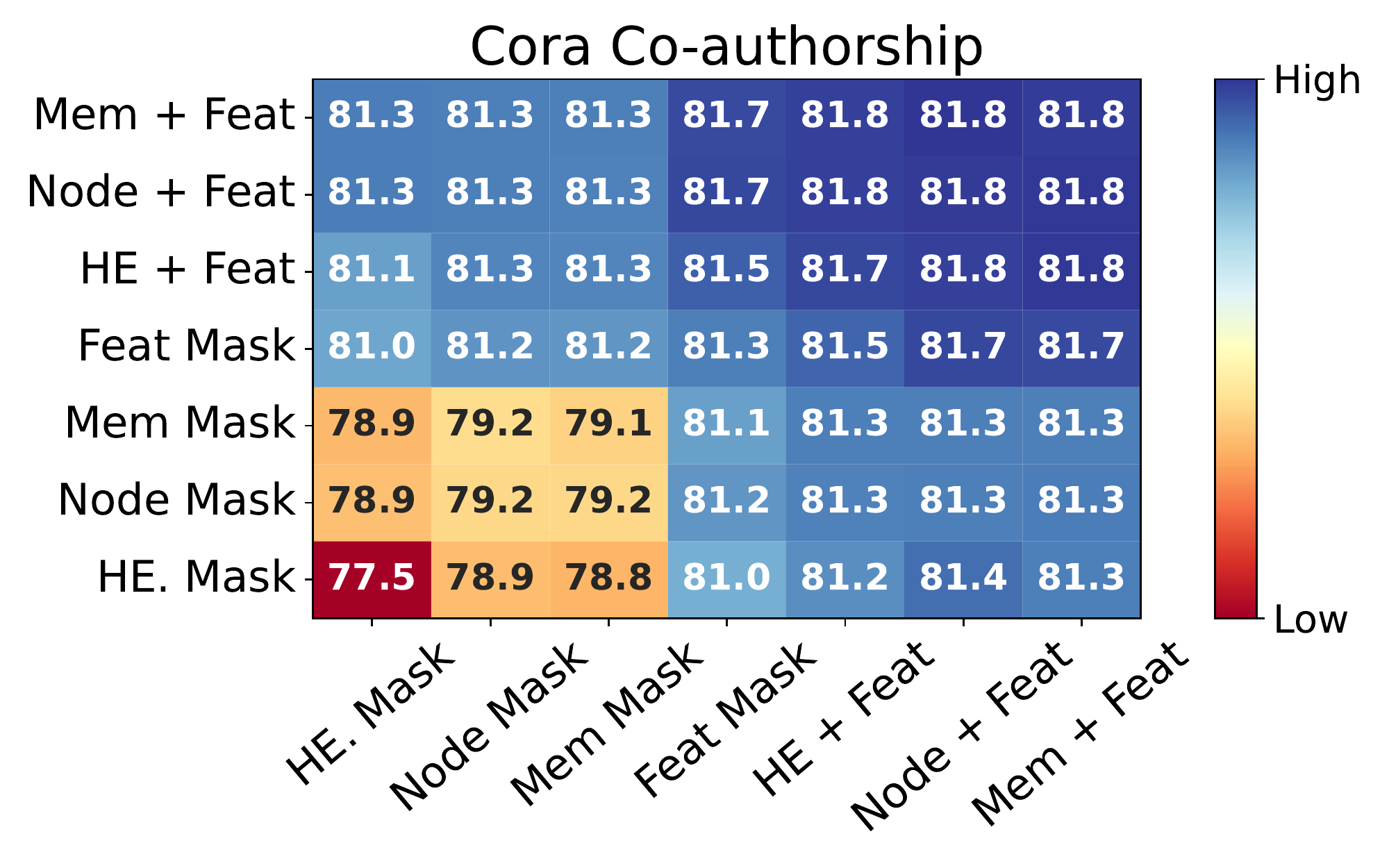}\\
    \vspace{-2mm}
    \includegraphics[width=0.8\linewidth]{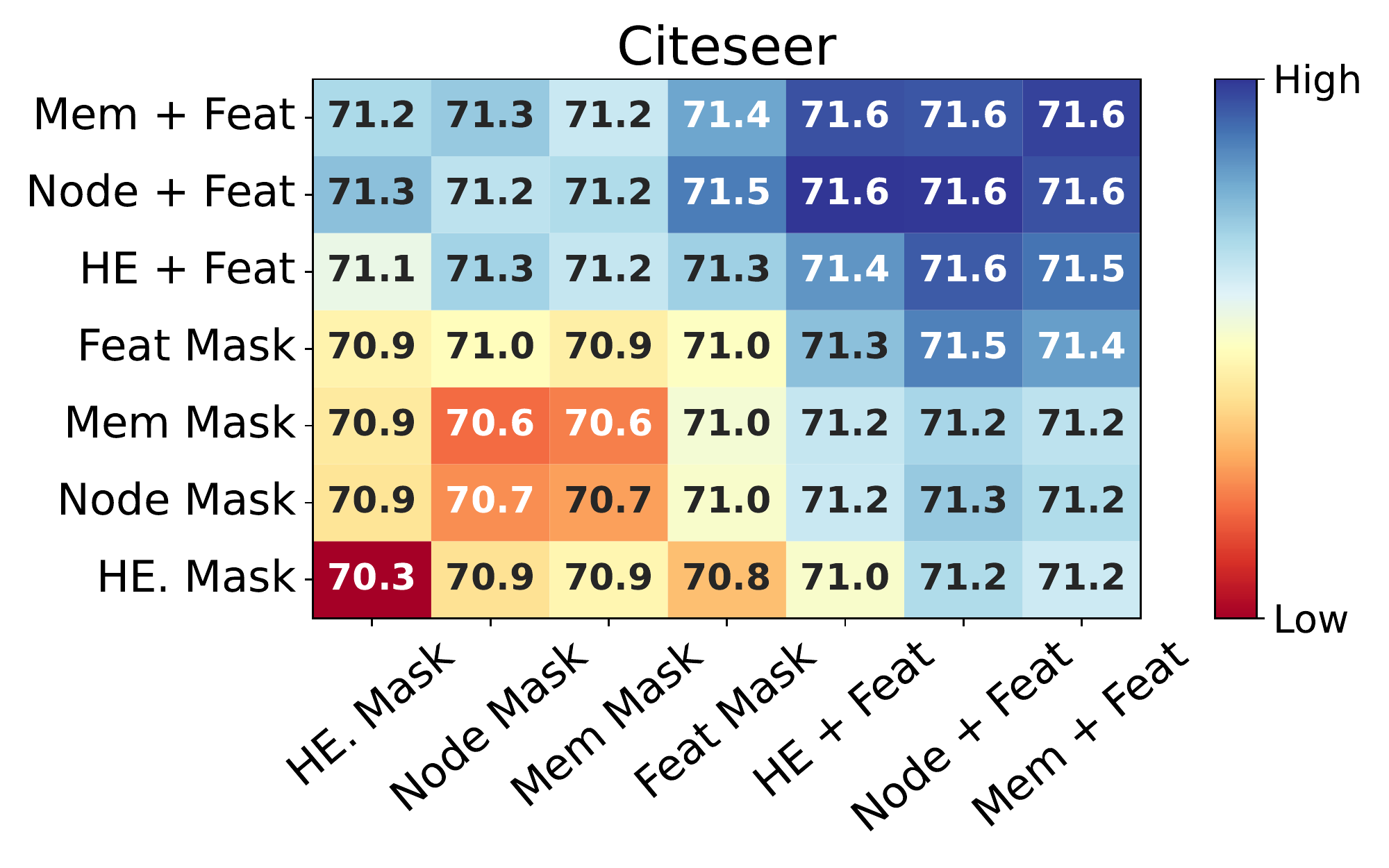}
    \vspace{-2mm}
    \caption{ \label{fig:app:augmentation_type}
    Node classification accuracy (\%) when employing different augmentation pairs.
    Using the structural and attribute augmentations together always yields better performance than using just one.
    }
    \vspace{-2mm}
\end{figure}

To show which types of augmentation are advantageous, we first examine the node classification performance for different augmentation pairs with a masking rate of 0.2.
We summarize the results in Figure~\ref{fig:app:augmentation_type}.
Note that, when using only one augmentation for each view, the effect of node feature masking is consistently good, but in particular, hyperedge masking performs poorly.
Next, using the structural and attribute augmentations together always yields better performance than using just one.
Among them, the pair of membership masking and node feature masking shows the best performance, demonstrating that using it in \method is a reasonable choice.
The combination of node masking and node feature masking is also a good choice.

\begin{figure}[!t]
    \centering
    \includegraphics[width=0.8\linewidth]{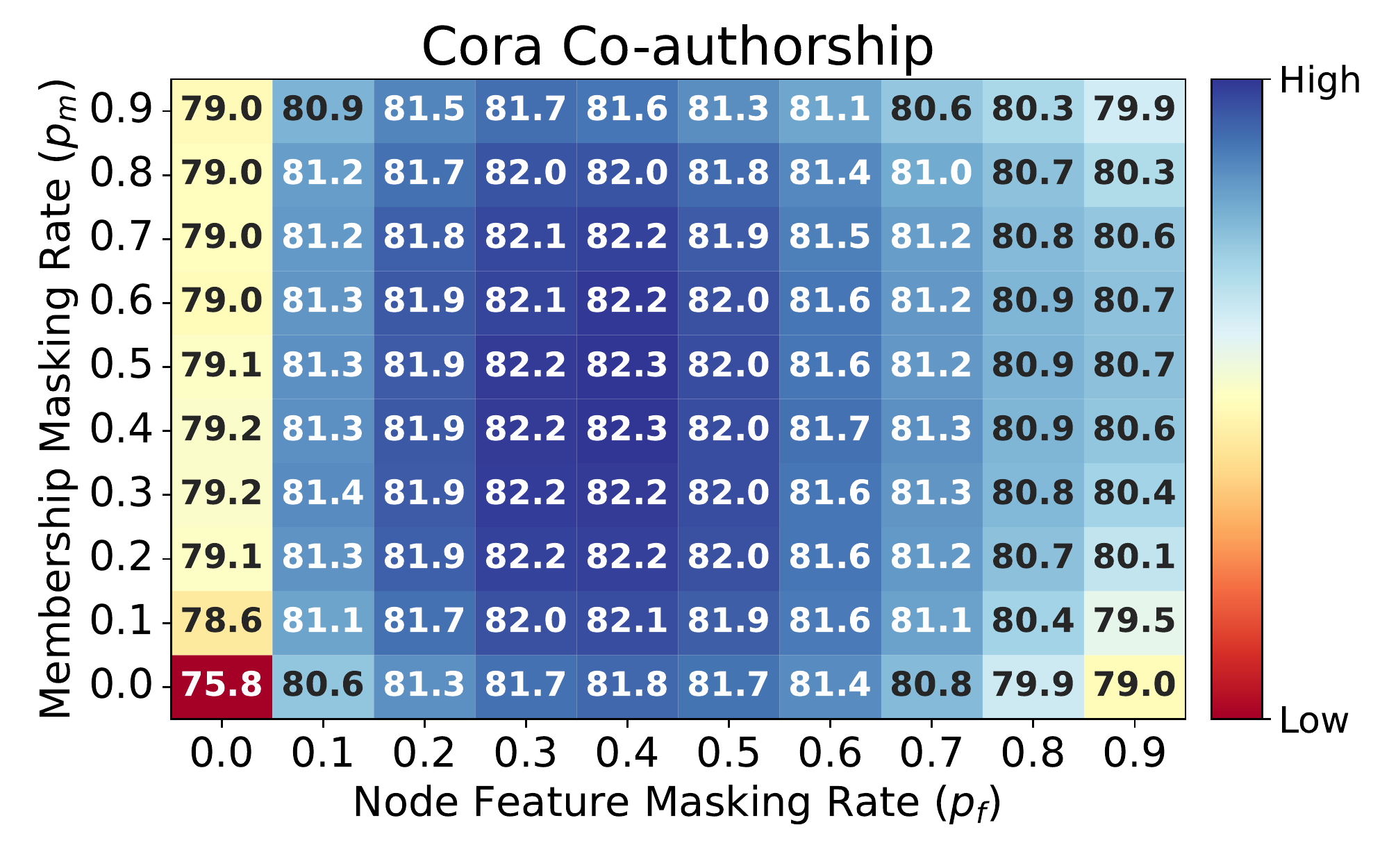}\\
    \vspace{-2mm}
    \includegraphics[width=0.8\linewidth]{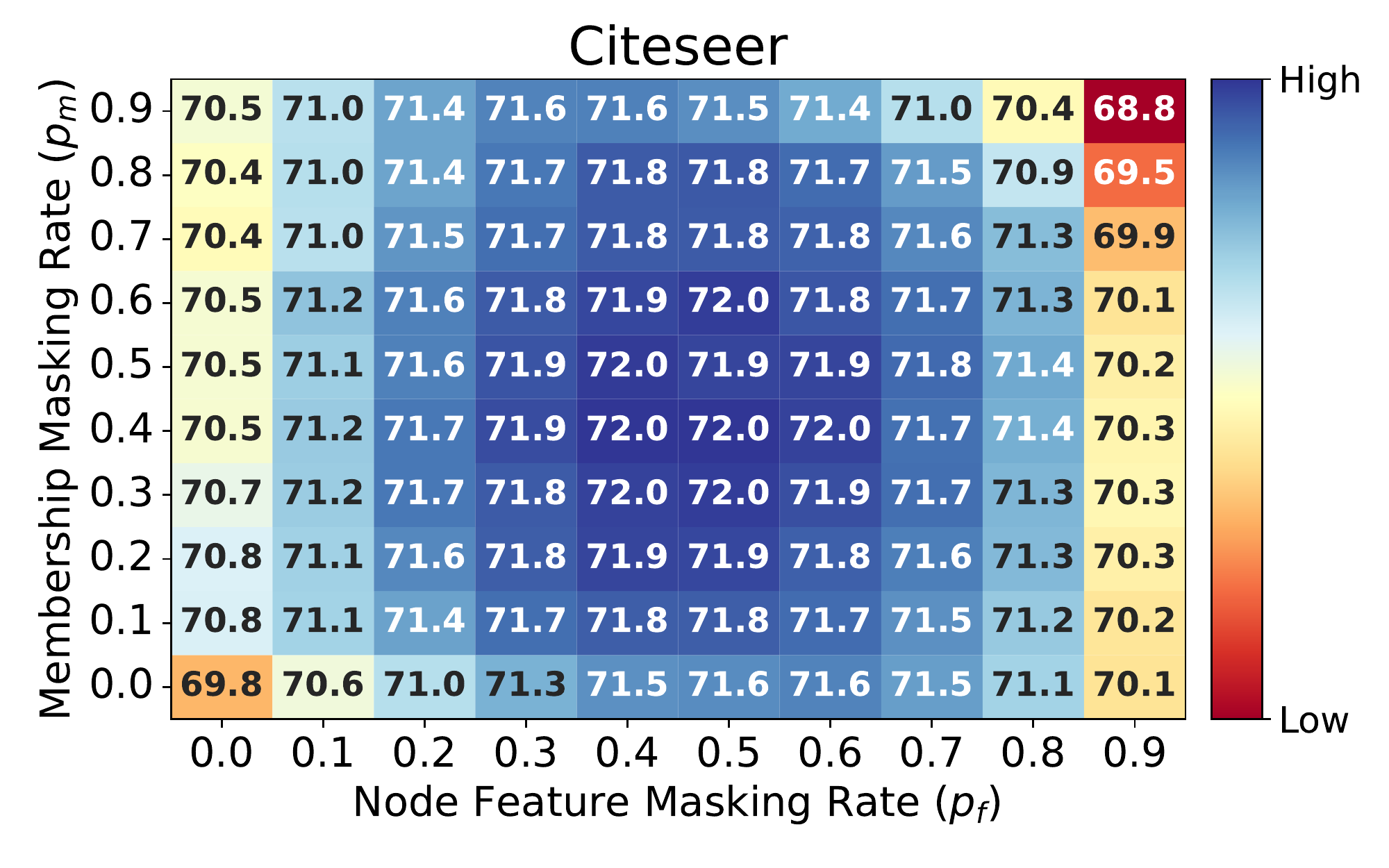}
    \vspace{-2mm}
    \caption{ \label{fig:app:augmentation_degree}
    Node classification accuracy (\%) according to the masking rates of membership and node features.
    A moderate extent of augmentation (i.e., masking rate between 0.3 and 0.7) benefits the downstream performance most.
    }
\end{figure}

Figure~\ref{fig:app:augmentation_degree} shows the node classification accuracy according to the membership and the node feature masking rate.
It demonstrates that a moderate extent of augmentation (i.e., masking rate between 0.3 and 0.7) benefits the downstream performance most.
If the masking rate is too small, two similar views are generated, which are insufficient to learn the discriminant ability of the encoder, and if it is too large, the underlying semantic of the original hypergraph is broken.

\section{Additional Experiments}
\label{sec:app:exp}

\subsection{Ablation Study: Effects of Self-loops}
\label{sec:app:exp:self}
In \method, we add self-loops to the hypergraph after hypergraph augmentation and before it is passed through the encoder.
We conduct an ablation study to demonstrate the effects of self-loops.
The results are summarized in Table~\ref{tab:selfloop}, and it empirically verifies that adding self-loops is advantageous. 
The reason for the better performance we speculate is that a self-loop helps each node make a better use of its initial features. 
Specifically, a hyperedge corresponding to a self-loop receives a message only from the 
node it contains and sends a message back to the node without aggregating the features of any other nodes. 
This allows each node to make a better use of its features.

\begin{table}[!t]
    \vspace{-2mm}
    \caption{
    Node classification accuracy (\%) of \method when the hypergraphs, which the encoder receives as input, are with and without self-loops.
    Adding self-loops helps to improve the quality of the node representations.
    }
    \label{tab:selfloop}
    \centering
    \scalebox{0.9}{
        \begin{tabular}{lcc}
            \toprule
            \multirow{2}{*}[-1mm]{Dataset} & \multicolumn{2}{c}{\method} \\
            \cmidrule(lr){2-3}
            & Without Self-loops & With Self-loops (Proposed) \\
            \midrule
            \midrule
            Cora-C      & 80.95 $\pm$ 1.29  & \textbf{81.57 $\pm$ 1.12} \\
            Citeseer    & 71.10 $\pm$ 1.09  & \textbf{72.02 $\pm$ 1.16} \\
            Pubmed      & 84.02 $\pm$ 0.56  & \textbf{84.26 $\pm$ 0.62} \\
            Cora-A      & 79.77 $\pm$ 0.95  & \textbf{82.15 $\pm$ 0.89} \\
            DBLP        & 90.19 $\pm$ 0.12  & \textbf{91.12 $\pm$ 0.11} \\
            Zoo         & \textbf{80.43 $\pm$ 11.3}  & 80.25 $\pm$ 11.2 \\
            20News      & 79.56 $\pm$ 0.24  & \textbf{80.14 $\pm$ 0.19} \\
            Mushroom    & 99.80 $\pm$ 0.14  & \textbf{99.83 $\pm$ 0.13} \\
            NTU2012     & 74.01 $\pm$ 2.54  & \textbf{75.23 $\pm$ 2.45} \\
            ModelNet40  & 93.48 $\pm$ 0.16  & \textbf{97.08 $\pm$ 0.13} \\
            \bottomrule
        \end{tabular}
    }
    \vspace{-2mm}
\end{table}

\subsection{Ablation Study: Backbone Encoder}
\label{sec:app:exp:encoder}
The superiority of the encoder used in \method over \HGNN is verified in Table~\ref{tab:backbone}. 
We compare the accuracy of two \method models that use (1) \HGNN and (2) the mean pooling layer (proposed), respectively, as an encoder.
\method with the mean pooling layer consistently and slightly outperforms the one with \HGNN as an encoder.
This result justifies our choice of using the mean pooling layer as our backbone encoder.

\subsection{Sensitivity Analysis}
\label{sec:app:exp:sensitivity}
We investigate the impact of hyperparameters used in \method, especially, $\tau_{g}$ and $\tau_{m}$ in Eq. (\ref{eq:loss:hyperedge:sum}) and (\ref{eq:loss:membership:sum}) as well as $\weightg$ and $\weightm$ in Eq. (\ref{eq:loss:final}), with the Citeseer and Cora Co-citation datasets.
We only change these hyperparameters in this analysis, and the others are fixed as provided in Appendix~\ref{sec:app:impl:param}.

\begin{figure}[!t]
    \centering
    \includegraphics[width=0.8\linewidth]{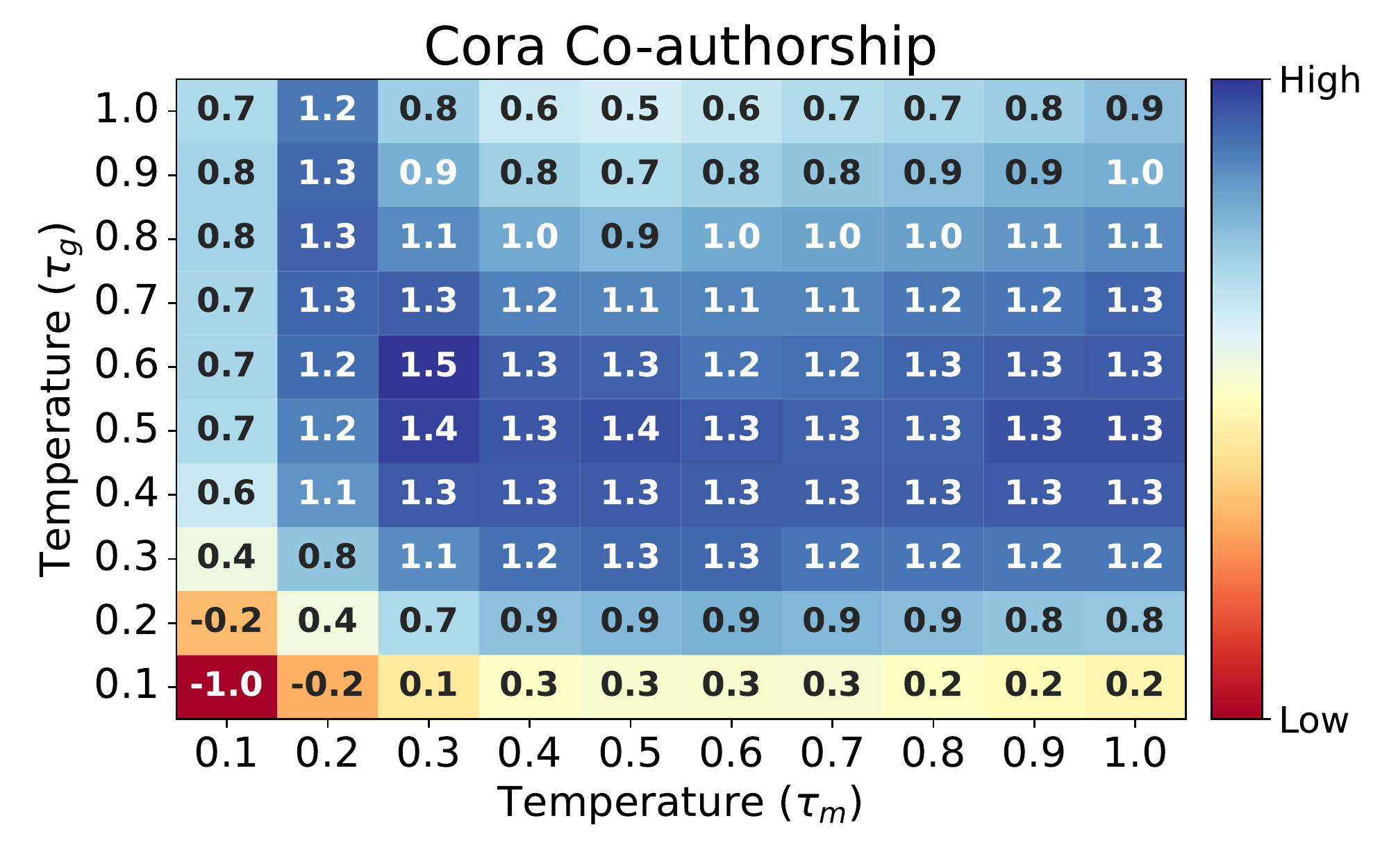}
    \vspace{-2mm}
    \includegraphics[width=0.8\linewidth]{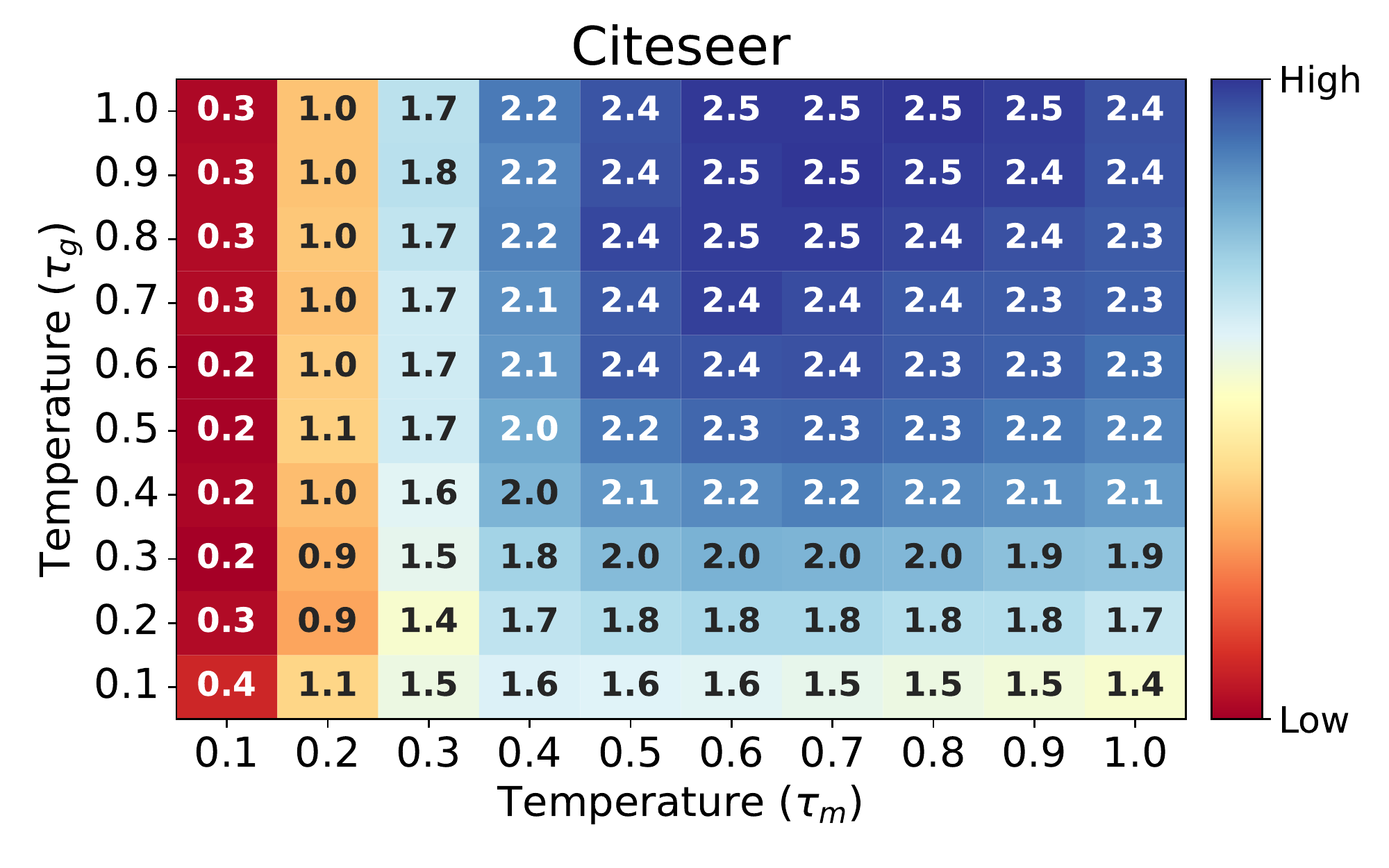}\\
    \vspace{-2mm}
    \caption{ \label{fig:app:sensitivity_tau}
    Node classification accuracy gain (\%) of \method over \methodN, when different temperature parameter pairs are used. 
    The baseline accuracies are 70.28\% and 80.49\% for the Citeseer and Cora Co-citation datasets, respectively.
    }
\end{figure}

\begin{figure}[!t]
    \centering
    \includegraphics[width=0.8\linewidth]{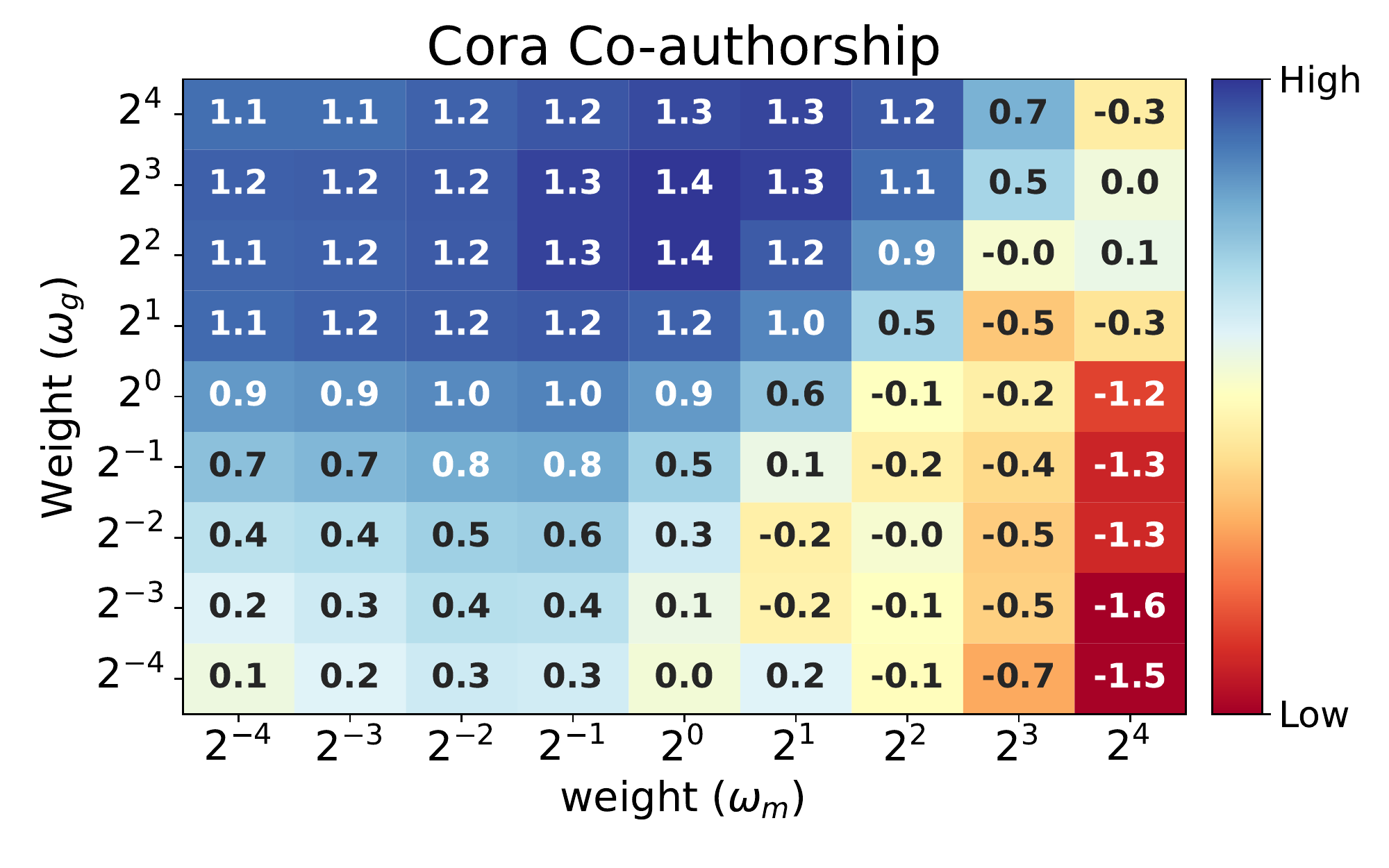}
    \vspace{-2mm}
    \includegraphics[width=0.8\linewidth]{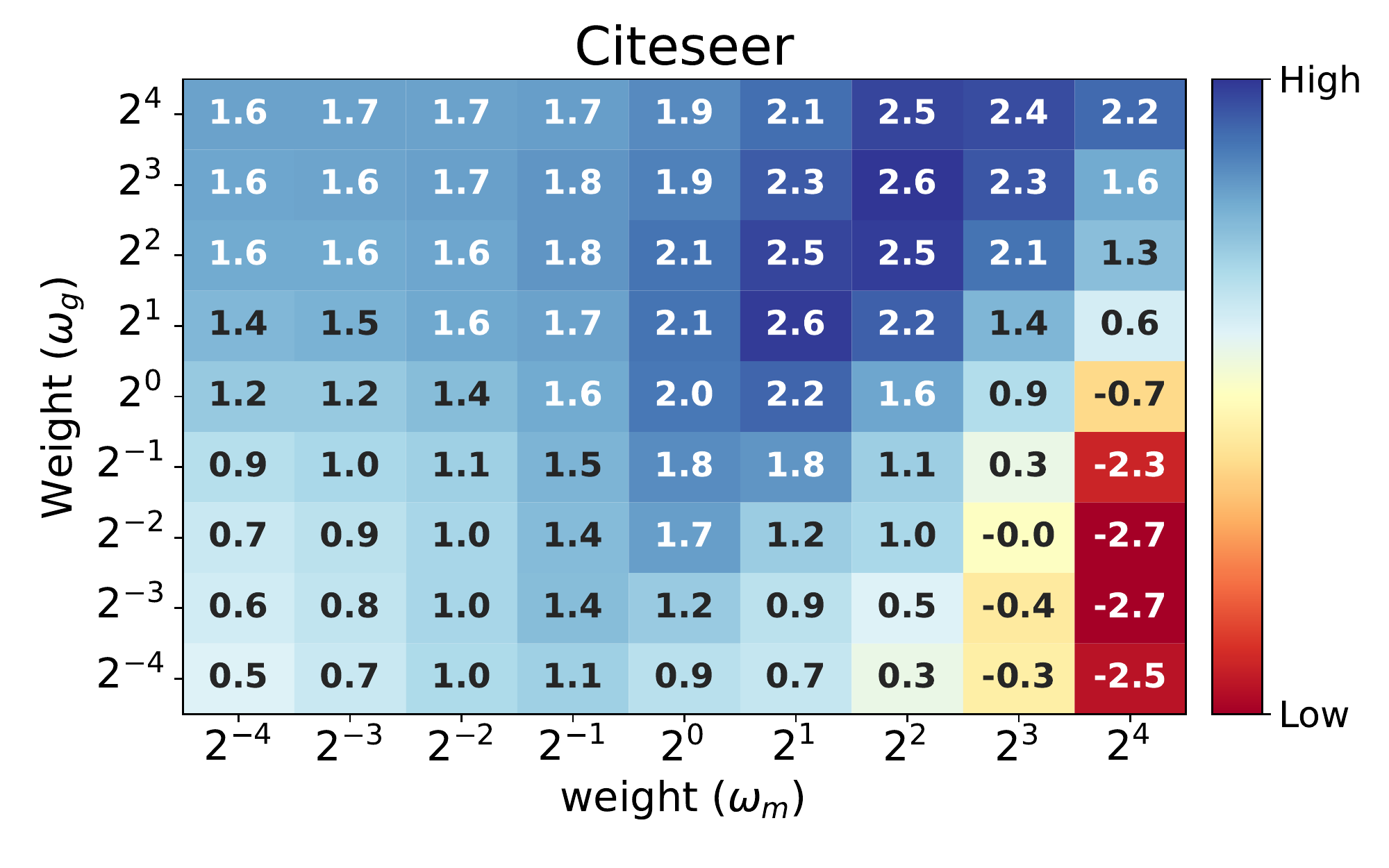}\\
    \vspace{-2mm}
    \caption{ \label{fig:app:sensitivity_w}
    Node classification accuracy gain (\%) of \method over \methodN, when different weight pairs are used.
    The baseline accuracies are 70.28\% and 80.49\% for the Citeseer and Cora Co-citation datasets, respectively.
    }
    \vspace{-2mm}
\end{figure}

We conduct node classification while varying the values of $\tau_{g}$ and $\tau_{m}$ from 0.1 to 1.0 and report the accuracy gain over \methodN, which only considers node-level contrast, in Figure~\ref{fig:app:sensitivity_tau}. 
From the figure, it can be observed that \method achieves an accuracy gain in most cases when both the temperature parameters are not too small (i.e., 0.1), as shown in the blue area in the figure.
It indicates that pursuing excessive uniformity in the embedding space rather degrades the node classification performance~\citep{wang2021understanding}.
We also conduct the same task while varying the values of $\weightg$ and $\weightm$ from $2^{-4}$ to $2^{4}$ and report the accuracy gain over \methodN, in Figure~\ref{fig:app:sensitivity_w}.
Using a large $\weightg$ and a small $\weightm$ together degrades the performance.
This causes model collapse by making the proportion of membership contrastive loss relatively larger than node and group contrastive losses.

\begin{table}[!t]
    \caption{
    The accuracy of two \method methods that use (1) \HGNN and (2) the mean pooling layer (proposed), respectively, as an encoder. 
    }
    \label{tab:backbone}
    \centering
    \scalebox{0.9}{
        \begin{tabular}{lcc}
            \toprule
            Dataset & \method (with \HGNN) & \method (Proposed) \\
            \midrule
            \midrule
            Cora-C      & 81.03 $\pm$ 1.31  & \textbf{81.57 $\pm$ 1.12} \\
            Citeseer    & 71.97 $\pm$ 1.26  & \textbf{72.02 $\pm$ 1.16} \\
            Pubmed      & 83.80 $\pm$ 0.63  & \textbf{84.26 $\pm$ 0.62} \\
            Cora-A      & \textbf{82.22 $\pm$ 1.09}  & 82.15 $\pm$ 0.89 \\
            DBLP        & 90.93 $\pm$ 0.16  & \textbf{91.12 $\pm$ 0.11} \\
            Zoo         & 79.47 $\pm$ 11.0  & \textbf{80.25 $\pm$ 11.2} \\
            20News      & 79.93 $\pm$ 0.24  & \textbf{80.14 $\pm$ 0.19} \\
            Mushroom    & 98.93 $\pm$ 0.33  & \textbf{99.83 $\pm$ 0.13} \\
            NTU2012     & 74.63 $\pm$ 2.53  & \textbf{75.23 $\pm$ 2.45} \\
            ModelNet40  & \textbf{97.33 $\pm$ 0.14}  & 97.08 $\pm$ 0.13 \\
            \bottomrule
        \end{tabular}
    }
    \vspace{-2mm}
\end{table}

\subsection{Training Time Comparison}
\label{sec:app:exp:time}

We compare the training time of baseline models and \method by the elapsed time of a single epoch. 
We run each method for 50 epochs and measured the average elapsed time per epoch (ms). 
Note that subsampling is not used and \method computes the membership contrastive loss with mini-batches of size 4096 when training for the Pubmed, DBLP, 20News, Mushroom, and ModelNet40 datasets due to the memory limits. 
Generally, \methodNE shows similar execution times to baseline approaches, and \method is slower due to membership contrast.

\begin{table*}[!t]
    \caption{Single epoch running time (in milliseconds) averaged over 50 training epochs. 
    The hyphen(-) indicates that the running time cannot be measured due to out of memory.}
    \label{tab:training_time}
    \centering
    \scalebox{0.8}{
        \begin{tabular}{lcccccccccc}
            \toprule
             & Cora-C & Citeseer & Pubmed & Cora-A & DBLP & Zoo & 20News & Mushroom & NTU2012 & ModelNet40 \\
            \midrule
            \midrule
                \DGI        & 4     & 4     & 16    & 4     & 76    & 4     & -   & -   & 4     & 13  \\
                \GRACE      & 15    & 15    & 32    & 16    & -   & 16    & -   & -   & 18    & 93  \\
                \HHGR       & 25    & 25    & 54    & 36    & 524   & 9     & 185   & 94    & 33    & 142  \\
            \midrule
                \methodN    & 13    & 12    & 18    & 11    & 488   & 10    & 78    & 37    & 11    & 50    \\
                \methodNE   & 19    & 17    & 38    & 17    & 625   & 15    & 83    & 40    & 16    & 85    \\
                \method     & 102   & 79    & 652   & 121   & 3,156 & 32    & 396   & 702   & 194   & 636   \\
            \bottomrule
        \end{tabular}
    }
\end{table*}
\section{Qualitative Analysis}
\label{sec:app:qualitative}

\subsection{Analysis on the Pubmed, Cora Co-authorship, and DBLP datasets}
\label{sec:app:qualitative:other}
As additional experiments, in Table~\ref{tab:app:tsne}, We provide t-SNE~\citep{van2008visualizing} plots of the node representations produced by \method and its two variants, \methodN and \methodNE, on the Pubmed, Cora Co-authorship, and DBLP datasets.
As expected from the quantitative results, the 2-D projection of embeddings learned by \method shows more numerically (based on Silhouette score~\citep{rousseeuw1987silhouettes}) distinguishable clusters than its two variants.

\begin{table*}[!t]
    \caption{
    t-SNE plots of the node representations from \method and its two variants.
    The \method's embeddings exhibits the most distinct clusters with the help of group and membership contrast, as measured by the Silhouette score (the higher, the better).
    }
    \label{tab:app:tsne}
    \centering
    \scalebox{0.80}{
        \begin{tabular}{cccc}
            \toprule
             & Pubmed & Cora Co-authorship & DBLP \\
            \midrule
            \midrule
            \rotatebox{90}{\hspace{13mm}\methodN} & 
            \hspace{-2mm} \includegraphics[width=0.27\linewidth]{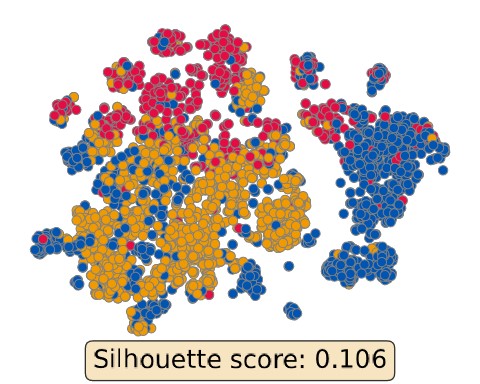} \hspace{-2mm} &
            \hspace{-2mm} \includegraphics[width=0.27\linewidth]{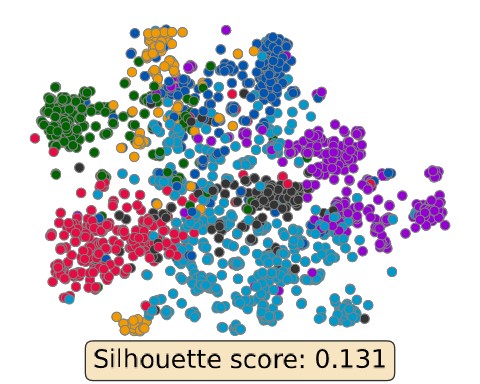} \hspace{-2mm} & 
            \hspace{-2mm} \includegraphics[width=0.27\linewidth]{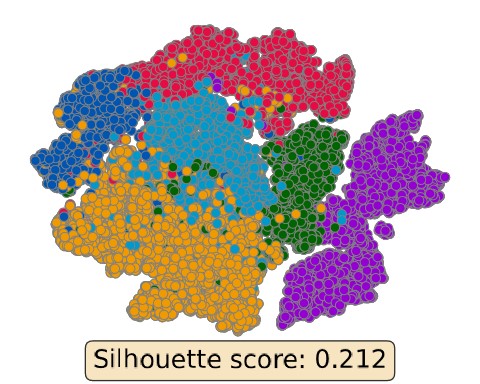} \hspace{-2mm} \\
            \midrule
            \rotatebox{90}{\hspace{11mm}\textbf{\methodNE}} & 
            \hspace{-2mm} \includegraphics[width=0.27\linewidth]{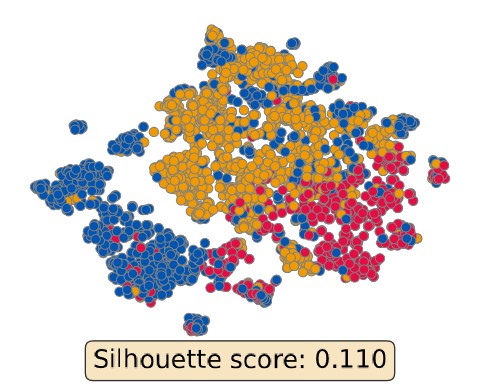} \hspace{-2mm} &
            \hspace{-2mm} \includegraphics[width=0.27\linewidth]{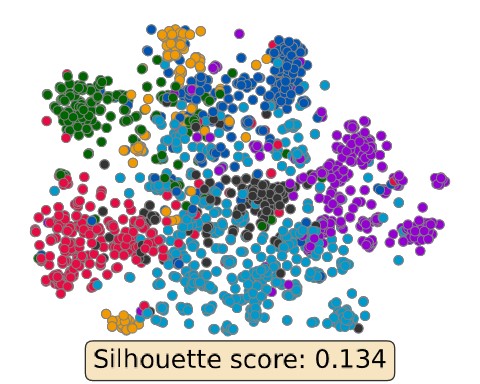} \hspace{-2mm} & 
            \hspace{-2mm} \includegraphics[width=0.27\linewidth]{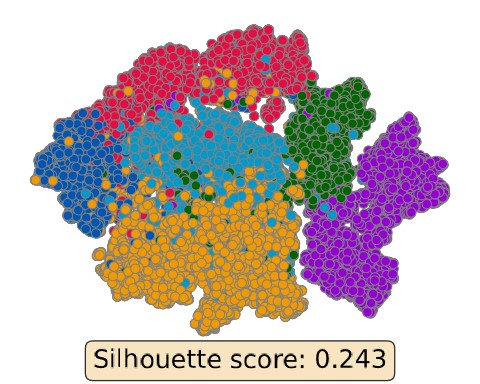} \hspace{-2mm} \\
            \midrule
            \rotatebox{90}{\hspace{15mm}\textbf{\method}} & 
            \hspace{-2mm} \includegraphics[width=0.27\linewidth]{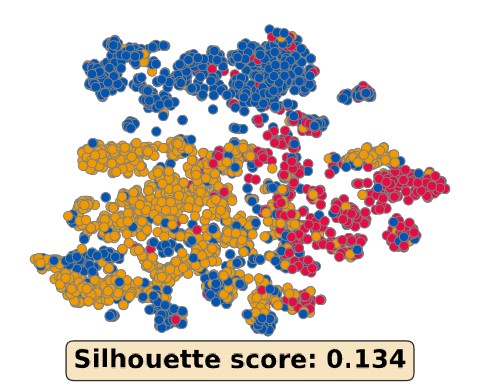} \hspace{-2mm} &
            \hspace{-2mm} \includegraphics[width=0.27\linewidth]{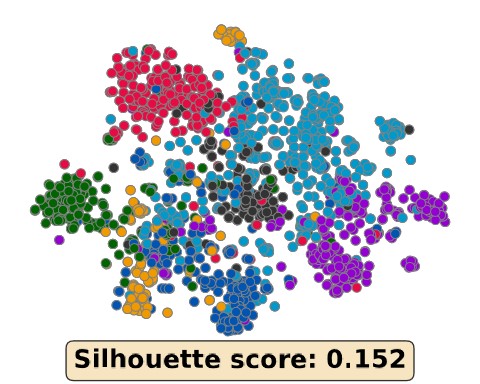} \hspace{-2mm} & 
            \hspace{-2mm} \includegraphics[width=0.27\linewidth]{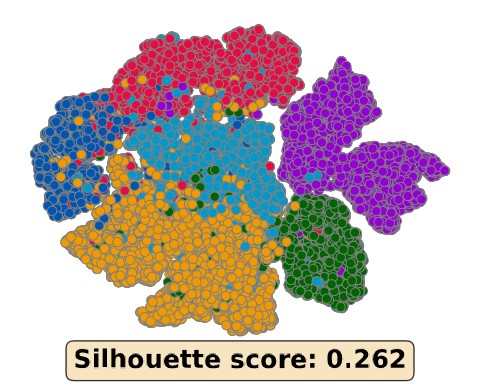} \hspace{-2mm} \\
            \bottomrule
        \end{tabular}
    }
\end{table*}

\subsection{Analysis on the Collapsed Models}
\label{sec:app:qualitative:collapse}

\begin{figure*}[!t]
    \centering
    \begin{subfigure}{0.25\textwidth}
        \centering
        \includegraphics[width=0.9\linewidth]{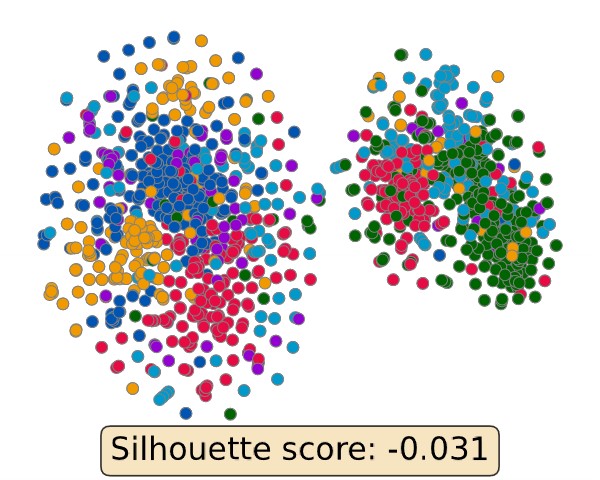}
        \caption{Citeseer}
    \end{subfigure}
    \begin{subfigure}{0.25\textwidth}
        \centering
        \includegraphics[width=0.9\linewidth]{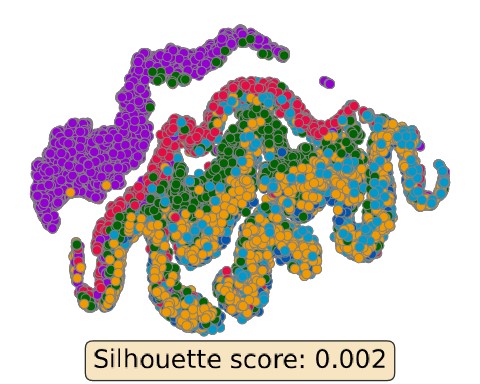}
        \caption{DBLP}
    \end{subfigure}
    \caption{ \label{fig:app:collapse}
    t-SNE plots of the node representations from \method when model collapse occurred. There is no clear distinction between the representations of nodes of different classes.
    }
\end{figure*}

Using membership contrast alone sometimes causes model collapse. 
t-SNE plots of the collapsed models are shown in Figure~\ref{fig:app:collapse}. 
There is no clear distinction between the representations of nodes of different classes, and they overlap.
It even looks randomly scattered around two clusters in the Citeseer dataset.
One potential reason the model fails to produce separable embeddings is that there is no guidance between node representations or between edge representations.
Using node or group contrast together, this problem could be solved.

\end{document}